\pdfoutput=1
\PassOptionsToPackage{backref=page}{hyperref}
\documentclass[11pt]{article}

\newif\ifarxiv
\arxivtrue %

\ifarxiv
    \usepackage{acl}
\else 
    \usepackage[review]{acl}
\fi

\usepackage{times}
\usepackage{latexsym}

\usepackage[T1]{fontenc}

\usepackage[utf8]{inputenc}

\usepackage[algo2e]{algorithm2e}

\usepackage{microtype}
\usepackage{sec/package}

\usepackage{multirow}
\usepackage{booktabs}
\usepackage{tabularx}

\usepackage{dsfont}

\usepackage{color}
\usepackage{soul}
\usepackage[textsize=scriptsize]{todonotes}

\title{

A Causal Framework to Quantify the Robustness of\\Mathematical Reasoning with Language Models

}

\author{Alessandro Stolfo\thanks{\hspace{0.07cm} Equal contribution.}
\\
  ETH Zürich \\
  \texttt{stolfoa@ethz.ch} \\\And
  Zhijing Jin\samethanks \\
  MPI \& ETH Zürich \\
  \texttt{jinzhi@ethz.ch} \\\AND
  Kumar Shridhar \\
  ETH Zürich \\
  \texttt{shkumar@ethz.ch} \\\And
  Bernhard Sch\"olkopf \\
  MPI \& ETH Zürich \\
  \texttt{bs@tue.mpg.de}\\\And
  Mrinmaya Sachan \\
  ETH Zürich \\
  \texttt{msachan@ethz.ch} \\
}

\begin{document}
\arxivtrue

\maketitle
\begin{abstract}
We have recently witnessed a number of impressive results on hard mathematical reasoning problems with language models. At the same time, the robustness of these models has also been called into question; recent works have shown that models can rely on shallow patterns in the problem description when generating a solution.
Building on the idea of behavioral testing, we propose a novel framework, which pins down the causal effect of various factors in the input, e.g., the surface form of the problem text, the operands, and math operators on the output solution.
By grounding the behavioral analysis in a causal graph describing an intuitive reasoning process, we study the behavior of language models in terms of robustness and sensitivity to direct interventions in the input space. We apply our framework on a test bed of math word problems.
Our analysis shows that robustness does not appear to continuously improve as a function of size, but 
the GPT-3 Davinci models (175B) achieve a dramatic improvement in both robustness and sensitivity compared to all other GPT variants.\footnote{Our code and data 
\ifarxiv
are available at \url{https://github.com/alestolfo/causal-math}.
\else
have been uploaded to the submission system, and will be open-sourced upon acceptance.
\fi
}
\end{abstract}

\section{Introduction}
\begin{figure}[t]
    \centering
    \includegraphics[width=\columnwidth]{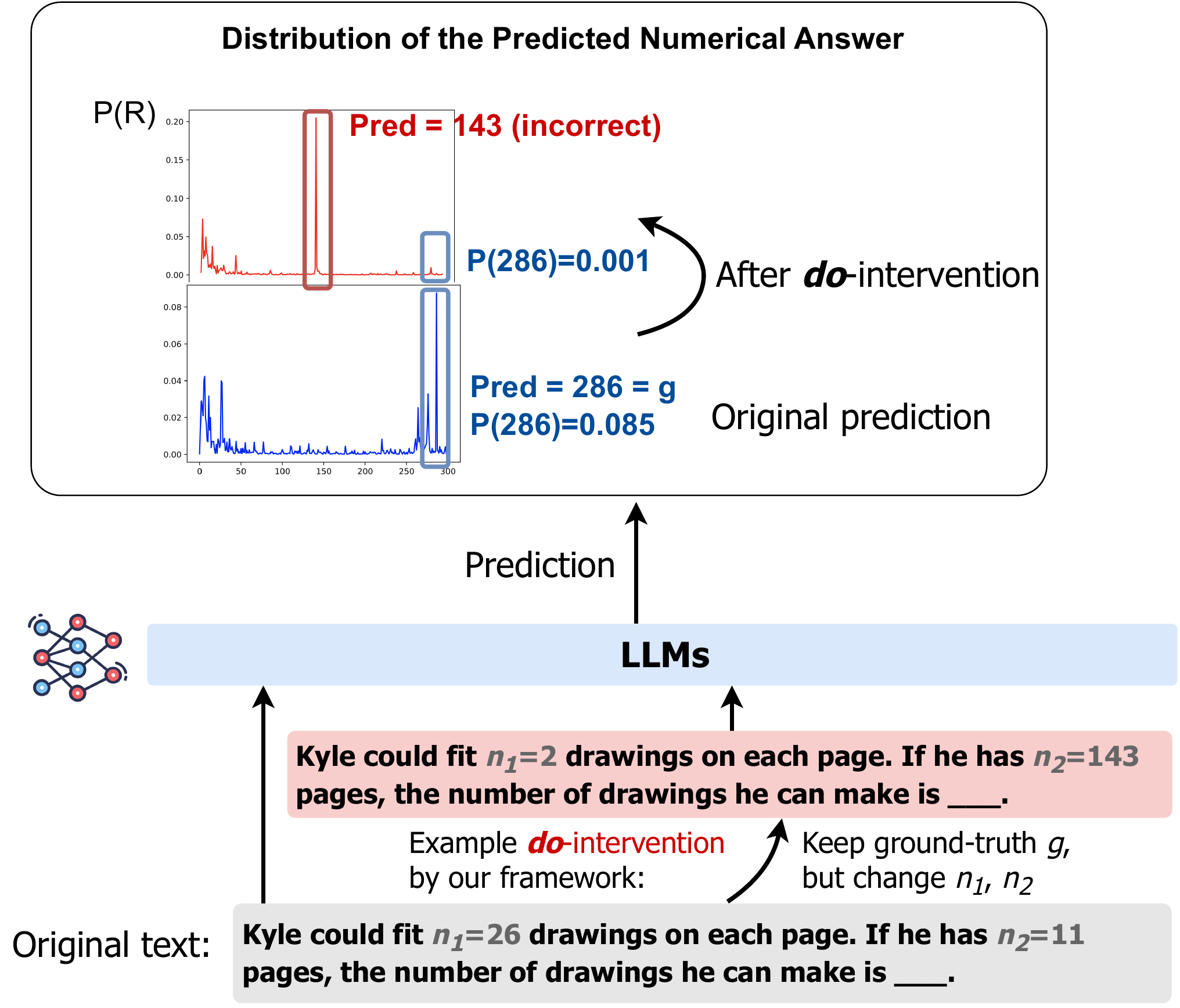}
    \caption{Through our framework, we conduct $\mathrm{do}$-interventions on the input and evaluate the change in the distribution $\mathbb{P}(R)$ of the prediction $R$ by LLMs, in this figure, GPT-J. This allows us to measure the causal effect of each factor in the input on the model's response.}
    \label{fig:intro}
    \vspace{-3mm}
\end{figure}
Many natural language understanding situations, such as understanding the financial news, require reasoning with text that includes numbers.
However, such mathematical reasoning is challenging for NLP models \cite{cobbe2021training, mishra-etal-2022-numglue}. %
Mathematical reasoning for text has been an active area of research for a while \citep[\textit{inter alia}]{seo-etal-2015-solving,sachan2017learning,sachan2017textbooks,sachan2018learning}, and has also emerged as a key task to track the capabilities of large language models (LLMs) in recent years \citep[\textit{inter alia}]{gpt3,ouyang2022instructGPT,wei2022emergent}.

\begin{figure*}[t]
    \centering
    \includegraphics[width=0.95\textwidth]{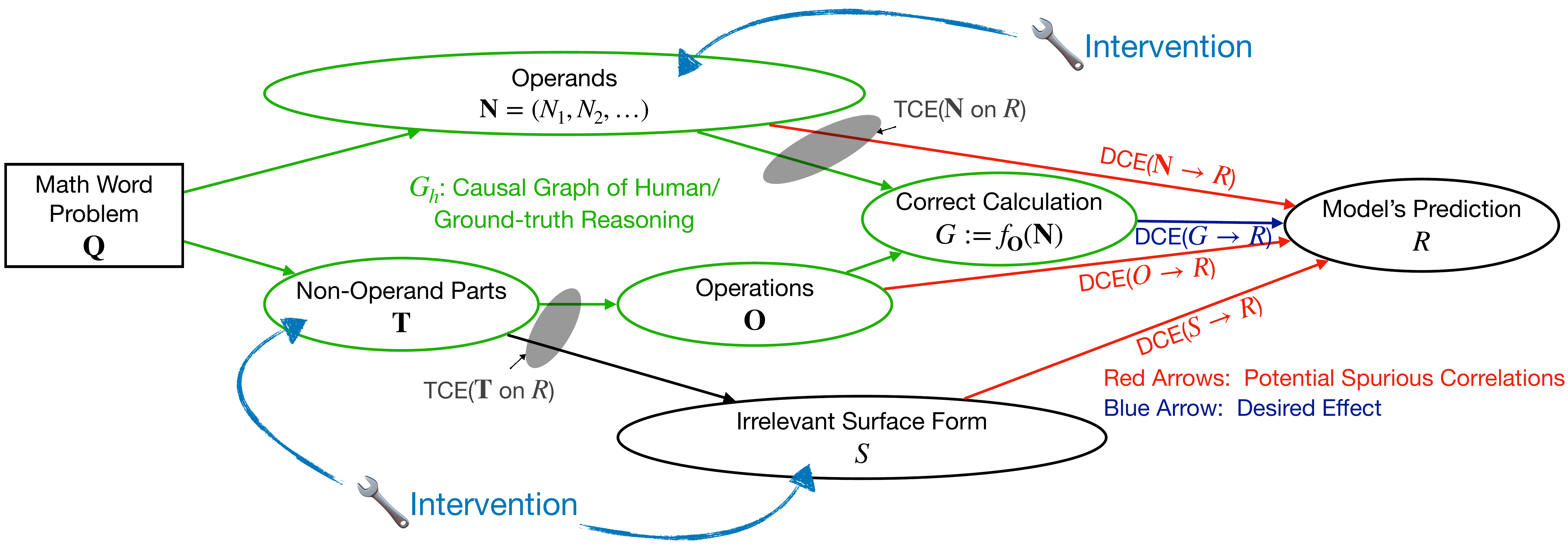}
    \caption{Causal graph of model predictions on math questions. We highlight the difference between a cognitively-inspired correct reasoning path (\green{$\mathcal{G}_h$}) and the undesired effects that some factors might have on the model's prediction (red arrows). By performing controlled interventions of the numerical values ($\bm{N}$) and on the textual framing of the problem ($\bm{T}$, $S$), we are able to quantify the causal effects of each factor.
    }
    \label{fig:causal_graph}
\end{figure*}

However, despite the impressive performance of LLMs on various math reasoning benchmarks \citep[e.g.,][]{ouyang2022instructGPT,chowdhery2022palm}, it remains unclear whether these models have %
learned mere artifacts in the data
or have truly mastered the mathematical concepts needed to consistently solve all variations of the same problem \citep{patel-etal-2021-nlp, razeghi2022impact,welleck2022symbolic}. In sharp contrast with a large number of papers on improving the performance of LLMs on various types of math-based problems, there has been little effort on behavioral analysis of LLMs for these tasks. Existing methods for understanding the robustness of these models \citep{patel-etal-2021-nlp} rely on manually constructing variations of math problems, and we do not yet have a principled, comprehensive framework for quantifying such robustness.%

Thus, in this work, we propose a formal framework based on causal inference, to quantify the robustness of NLP models' math reasoning abilities. Specifically, we describe a causal graph formulation of math reasoning, where the graph allows us to measure the difference in the structural causal models of human reasoning and model judgment. We consider various causal factors such as the textual framing of the question, numerical operands, and operation types. Then, we identify a set of interventions  in the context of math word problems (an example of which is illustrated in Figure \ref{fig:intro}), and provide a causal inference framework to obtain causal effects of each factor via direct $\mathrm{do}$-interventions \cite{pearl1995causal} and causal mediation analysis \cite{pearl2001direct}. While our approach is reminiscent of recent studies using causal analysis for LLMs \cite{finlayson-etal-2021-causal,vig2020investigating,meng2022locating}, in this work, we provide a new theoretical analysis framework specifically suitable for math reasoning.
Using our framework, we disentangle factors affecting the model's predictions and measure their influences. This way, we are able to provide insights into the model’s reasoning in terms of \emph{robustness} and \emph{sensitivity} with respect to changes in these factors.

We apply our framework to study a set of thirteen GPT models with various sizes and training procedures (i.e., instruction-tuned and non-instruction-tuned). 
We observe that, among non-instruction-tuned language models, the larger ones tend to be more sensitive to changes in the ground-truth result of a math word problem, but not necessarily more robust. 
However, we observe a different behavior in the instruction-tuned GPT-3 models \cite{ouyang2022instructGPT}, which show a remarkable improvement in both sensitivity and robustness, although the robustness reduces when problems get more complicated.
We additionally investigate the role of size and instruction tuning on the model's performance with three models of the LLaMA family \cite{touvron2023llama} and Stanford Alpaca \cite{alpaca}.

\section{Problem Setup} 
\label{sec:problem_setup}

We consider a dataset $\mathcal{D}$ of math word problems (MWPs), where each MWP is denoted as a question $\bm{Q}$. $\bm{Q}$ is a list $(\bm{T}, \bm{N})$ consisting of a question template $\bm{T}$ and an ordered list of operands $\bm{N} = (N_1, N_2, \dots, N_m)$. Each question template $\bm{T} := (\bm{O}, S)$ further contains two types of information: a set of arithmetic operations $\bm{O}$ implicitly expressed in the question, and the text surface form $S$ irrelevant to the arithmetic operations. $\bm{O}$ incorporates the information relative to the operations as a collection of tuples $\{(O_1, i_1, j_1), (O_2, i_2, j_2), \dots\}$, where $O_k \in \{+,-,\times,\div \}$ ($k \in \mathbb{N}$) and $i_k, j_k \in \mathbb{N}$ represent the indices of the operands to which operator $O_k$ should be applied to.\footnote{The intermediate result of operation $O_l$ is indicated by $i_k = m + l$.}
The ground-truth result $G = f_{\bm{O}}(\bm{N})$ is calculated by computing the function $f_{\bm{O}}$, which represents the application of all the operators in $\bm{O}$ to the respective operands.
We illustrate the factors in $\bm{Q}$ and their inter-dependency in the causal graph in Figure \ref{fig:causal_graph}.
A two-operand instance $\bm{q}$ of $\bm{Q}$ in this form from \citet{patel-etal-2021-nlp} is:
\vspace{-2mm}
\begin{quote}
    \textbf{Template $\bm{t}$}: Mark has $n_1$ trees in his backyard. If he plants $n_2$ more,
    how many trees will he have? \\
    \textbf{Operands $\bm{n}$}: $(n_1 = 12, n_2 = 13)$ \\
    \textbf{Operations} $\bm{o}$: \{(``$+$'', 1, 2)\} \\
    \textbf{Result}: $g= f_{\bm{o}}(\bm{n}) = n_1 + n_2 = 25$ \\
\end{quote}%
\vspace{-6mm}

Our goal is to quantify the robustness of a model $\mathcal{M}$ on the set of problems $\bm{q} \in \mathcal{D}$. Ideally, $\mathcal{D}$ should be a dataset not seen by the model during training.
We assume that a model takes $\bm{q}$ as input and predicts a probability distribution of the result $R$: $\mathbb{P}(R \ |\ \bm{t}, \bm{n})$.
Our formulation below will be easier to understand using this finite discrete set and can be generalized to any kind of data pairing a natural language template with a function that maps a set of operands to a result (e.g., a Python program; \citealt{mishra2022lila}).

\section{A Causal Framework}
In this section, we describe our framework in three steps. First, we define the idea of model robustness on MWPs.
Then, we identify possible $\mathrm{do}$-interventions \cite{pearl1995causal} that we can perform. Finally, we describe the causal effects that we measure to quantify the robustness of various models.

\subsection{Step 1. Question Reformulation}
We address the research question ``\textit{Is a model reasoning robustly on MWPs?}'' by comparing the causal mechanisms of the model's decisions to a hypothesized human reasoning mechanism.
Note that we do not claim to know how humans reason about these problems. We simply propose a reasonable and intuitive way to judge model robustness given a reasonable and intuitive human reasoning mechanism inspired by findings regarding the independence of language and mathematical reasoning in humans \cite{doi:10.1073/pnas.0500328102, monti2012thought}.

\paragraph{Human Reasoning Mechanisms.} The causal mechanisms of how humans might solve $\bm{q}$ include
\begin{align}
    \bm{o} & = f_{\mathrm{abstract}}(\bm{q})
    ~,
    \\
    g & = f_{\bm{o}}(\bm{n})
    ~,
\end{align}
where they first abstract the arithmetic operations $\bm{o}$ from the problem $\bm{q}$ by some cognitive process $f_{\mathrm{abstract}}$, and then apply the operation to the operands to obtain the result $g$. We show these mechanisms in the green subgraph \green{$\mathcal{G}_h$} of \cref{fig:causal_graph}.

\paragraph{Model Reasoning Mechanisms.} In contrast, the causal mechanisms of how a model might solve $\bm{q}$ are as follows:
\begin{align}
    r = f_{\mathrm{blackBox}} (\bm{t}, \bm{n})
    ~,
\end{align}
where we are unsure about (1) \textit{what} part(s) of $\bm{t}$ the model takes into account, and (2) \textit{how} it operates over the relevant variables.

Thus, we draw all possible causal mechanisms that might take place in the black-box model $f_{\mathrm{blackBox}}$ in the complete causal graph in \cref{fig:causal_graph}. Some possible fine-grained causal mechanisms are
\begin{enumerate}[nolistsep]
    \item The model might attend over the question template $\bm{t}$ in two ways: paying attention to the text surface form $s$ via the causal path $\bm{T} \rightarrow S \rightarrow R$, or text relevant to the math operations $\bm{o}$ via the causal path $\bm{T} \rightarrow \bm{O} \rightarrow R$.
    \item The model might also attend to the operands $\bm{n} := (n_1, n_2, \dots)$ via a causal path $\bm{N} \rightarrow R$.
    \item If the model learns the correct causal mechanisms as in the human cognitive process, it should capture how the operator and the operands matter to the ground-truth result $g$ (via $\bm{O} \rightarrow G$ and $\bm{N} \rightarrow G$) and then the model prediction should be sensitive to any changes in the ground truth, namely $G \rightarrow R$. No spurious correlations can directly affect $R$ without going through the mediator $G$.
\end{enumerate}

Hence, to answer the question ``How robust is the mathematical reasoning of a model on MWPs?'' we can answer the following subquestions:
\begin{enumerate}
    \item How does $R$ change in response to $G$? By quantifying this, we assess the \emph{sensitivity} (correct responsiveness) of the model to changes in the problem. In other words, does the model correctly adjust its prediction in response to a change in the correct solution of the problem?
    \item What is the (unwanted) direct causal effect size of $S \rightarrow R$,  and $\bm{N} \rightarrow R$? We see the quantities as a measure of the \emph{brittleness} (i.e., wrong responsiveness) of the model to result-preserving changes in the input. The lower the direct causal effect of $S$ and $\bm{N}$, the more \emph{robust} the model is.
\end{enumerate}

\subsection{Step 2. Causal Intervention List}\label{sec:quantities_of_interest}
After formulating the cognitively-inspired subgraph \green{$\mathcal{G}_h$} and defining the undesired causal paths in Figure \ref{fig:causal_graph}, we list all feasible limited actions that allow us to perform our causal analysis. In the context of MWPs, we use the following interventions:
\begin{enumerate}
    \item Direct intervention on all possible $n_1, n_2, \dots$;
    \item Partially controllable interventions on $\bm{T}$. We can replace the template $\bm{T}$ in two ways:
    \begin{enumerate}
        \item both $S$ and $\bm{O}$ are affected, or
        \item $S$ is affected but $\bm{O}$ is not affected.
    \end{enumerate}
\end{enumerate}

\subsection{Step 3. Turning Limited Actions into Causal Effect Sizes}
Next, we explain how we can obtain the causal effect sizes we want (listed in Step 1) from the limited set of interventions we can do (listed in Step 2).
Specifically, we first start from all the feasible interventions, and for variables that we cannot directly intervene on, we 
apply deductions from $\mathrm{do}$-calculus~\cite{pearl1995causal} to obtain or approximate the direct causal effect sizes. 
In the following, we describe a list of causal effect sizes that we need.

\paragraph{General Formulation.}

Let us consider an intervention $\mathrm{do}(X: x \rightarrow x')$, where $X \in \{\bm{T}, S, \bm{N}\}$ and a problem $\bm{Q} = \{\bm{T}, \bm{N}\}$.
The support of the numerical values $N_i$'s and $R$ is $\mathcal{I} \subseteq \mathbb{N}$, and we consider $\bm{N}$ to be distributed uniformly over the set $\{ \bm{n} \in \mathcal{I}^2 \ | \ f_{\bm{O}}(\bm{n}) \in \mathcal{I}\}$.
We denote the distribution before the intervention $\mathbb{P}(R \ | \ \bm{T}, \bm{N})$ as $P$ and the distribution after the intervention
as $P'$. 

Following the distributional definition of causal effect by \citet{pearl1995causal}, 
we quantify the effect of factor $X$ in our causal graph using a distance metric $\delta$ between the distributions $P$ and $P'$. That is, 
\begin{align}
    \mathrm{CE} = \delta(P, P'),
\end{align}

where $\mathrm{CE}$ can refer to the \textbf{total causal effect} (TCE, i.e., the joint effect through all the directed causal paths from a variable to another), or the \textbf{direct causal effect} (DCE, i.e., the effect from the directed causal path from a variable to another that does not go through any intermediate variables) \cite{pearl2001direct}.
We describe our choices for $\delta$ in Section \ref{sec:distr_diff_metrics}. 

\paragraph{Causal Effects of the Operands.}
When intervening on the operands $\bm{N}:=(N_1, N_2, \dots)$, we can obtain the size of the total causal effect of $\bm{N}$ on $R$, namely
\begin{align}
    \label{eq:tce_n_r}
     & \mathrm{TCE}(\bm{N} \text{ on } R) 
    := \mathbb{E}_{\bm{n}'\sim \mathbb{P}(\bm{N})} [\delta(P, P')],\\
    & \text{where } P' = \mathbb{P}(R|\bm{T}, \mathrm{do}(\bm{N} = \bm{n}'))
    ~.
\end{align}
Note that this TCE is not the exact desired quantity, because we want to separate two different paths of how $\bm{N}$ affects $R$: (1) the path $\bm{N} \rightarrow G \rightarrow R$, which is the correct decision path that we want the model to pick up (where the model reacts to the change in the ground-truth answer), and (2) the path $\bm{N} \rightarrow R$, which is the spurious correlation that the model might have learned (where the model relies on some spurious correlations with certain numerical values, which could be traced to perhaps their frequencies in the training corpus).

We can quantify the \textbf{direct causal effect} (DCE, i.e., the effect from the directed causal path from a variable to another that does not go through any intermediate variables) \cite{pearl2001direct} of $\bm{N}$ on $R$, namely the strength of the direct causal path $\bm{N} \rightarrow R$, by controlling for $G$ to be fixed every time we intervene on $\bm{N}$:
\begin{align}
    \label{eq:dce_n_r}
     & \mathrm{DCE}(\bm{N} \rightarrow R) 
    := \mathbb{E}_{\bm{n}' \sim \mathbb{P}(\bm{N} | G)} [\delta(P, P')],\\
    & \text{where } P' = \mathbb{P}(R|\bm{T}, \mathrm{do}(\bm{N} = \bm{n}'))
    ~.
\end{align}
For example, if we observe a model doing $100+100=200$ correctly, we want to separate the math ability here into (1) the model's sensitivity towards the ground-truth answer, and (2) the model's decisions based on its familiarity with just the operand $100$.
Here, the overall effect is the calculable $\mathrm{TCE}(\bm{N} \text{ on } R) $ by Eq. \ref{eq:tce_n_r}, and one of the subeffects is the calculable $\mathrm{DCE}(\bm{N} \rightarrow R)$ by Eq. \ref{eq:dce_n_r}.

\paragraph{Causal Effects of the Text Surface Form.}

As for the operands, we can compute both the direct and indirect effects of the surface form representing the math problem. In particular, intervening on $\bm{T}$ without controlling for $\bm{O}$ (intervention 2a in Sec. \ref{sec:quantities_of_interest}), we can compute the total effect, i.e.,
\begin{align}
     & \mathrm{TCE}(\bm{T} \text{ on } R) 
    := \mathbb{E}_{\bm{t}'\sim \mathbb{P}(\bm{T})} [\delta(P, P')],\\
    & \text{where } P' = \mathbb{P}(R|\bm{N}, \mathrm{do}(\bm{T} = \bm{t}'))
    ~.
\end{align}

Controlling for the operations $\bm{O}$ (intervention 2b in Sec. \ref{sec:quantities_of_interest}) will instead allow us to obtain the direct causal effect of the surface text:
\begin{align}
     & \mathrm{DCE}(S \rightarrow R) 
    := \mathbb{E}_{\bm{t}'\sim \mathbb{P}(\bm{T} | O)} [\delta(P, P')],\\
    & \text{where } P' = \mathbb{P}(R|\bm{N}, \mathrm{do}(\bm{T} = \bm{t}'))
    ~.
\end{align}
Note that since there is no mediator between $S$ and $R$, the $\mathrm{DCE}(S \rightarrow R)$ is also TCE of $S$ on $R$.
The only adaptation that we need to make with regard to the MWPs is that it is not feasible to enumerate all possible perturbations of $S$. Therefore, the practical results that researchers can achieve are over a certain subset of $S$. In practice, we obtain this by intervening on $\bm{T}$ without affecting $\bm{O}$.

\paragraph{Causal Effects of the Operators.}

The ideal way to obtain the TCE of $\bm{O}$ on $R$ is through some careful human annotation that minimally changes the templates as \citet{kaushik2020learning} do for sentiment classification. 
The challenge for MWPs in our case is that with all our possible interventions, we cannot \textit{only} intervene on $\bm{O}$ without introducing changes to the irrelevant surface form.
However, we might get some information about
$\mathrm{TCE}(\bm{O} \text{ on } R)$ because, on the causal graph, the total causal influence of $\bm{T}$ on $R$ actually flows into two directed paths, one through $S$ to $R$ (which is the $\mathrm{DCE}(S \rightarrow R)$), and the other from $\bm{O}$ to $R$, which is our interested quantity $\mathrm{TCE}(\bm{O} \text{ on } R)$. Therefore, we compare the two quantities we know, $\mathrm{TCE}(\bm{T} \rightarrow R)$ and $\mathrm{DCE}(S \rightarrow R)$, to get a sense of the causal influence of $\bm{O}$ on $R$ that we cannot obtain in any other way.

\subsection{Step 4. Quantifying the Causal Influence}\label{sec:distr_diff_metrics}

Consider a realization of problem $\bm{Q}$ with operands $\bm{n}$ and ground-truth result $g = f_{\bm{o}}(\bm{n})$, and denote by $g'$ the result after the intervention $\mathrm{do}(X: x \rightarrow x')$.
We quantify the causal effect of factor $X$ on the model's prediction $R$ in two ways: by assessing the change in the predicted result, and by measuring the change in the probability assigned by the model to the correct result $g$ (or $g'$). 

\myparagraph{Change in the Prediction}
To account for the inability of LMs to capture the continuous property of numbers \cite{jin2021numgpt}, we measure the change in the model's prediction using an indicator of the ``change result'' event:
\begin{align}
    \delta_{\mathrm{cp}} ({P}, {P'}) := \mathds{1} (r \neq r')
    ~,
\end{align}
where $r = \argmax_{x \in \mathcal{I}}P(x)$, and $r' = \argmax_{x \in \mathcal{I}}P'(x)$.

\myparagraph{Relative Change in Confidence}
Inspired by \citet{finlayson-etal-2021-causal}, we also highlight the change in terms of the relative difference in the probability assigned to $g$ and $g'$. We formulate two types of relative change, one quantifying the relative change in the confidence of $g$, and the other quantifying the relative change in the confidence of $g'$:
\begin{align}
    \Delta_{\mathrm{rel}} &= \frac{ {P}(g)- {P'}(g) }{{P'}(g)} \\
    \Delta_{\mathrm{rel}}'& = \frac{ {P'}(g')- {P}(g') }{{P}(g')} ~.
\end{align}

We quantify the overall relative change in confidence (RCC) as the average of the two relative changes above:
\begin{align}
     \delta_{\mathrm{rcc}} ({P}, {P'}) = \frac{1}{2} \bigg(\Delta_{\mathrm{rel}}  +  \Delta_{\mathrm{rel}}' \bigg)~.
\end{align}

\myparagraph{A Unified Form}
We are interested in the average causal effect of the intervention across all problems in $\mathcal{D}$. Thus, we measure the average of the effects over all instances $\bm{q} \in \mathcal{D}$.
We denote by the subscripts $\mathrm{TCE}_{\mathrm{cp}}$/$\mathrm{DCE}_{\mathrm{cp}}$ and $\mathrm{TCE}_{\mathrm{rcc}}$/$\mathrm{DCE}_{\mathrm{rcc}}$ the causal effects computed using the change in prediction metric and the relative change in confidence, respectively.
We describe how we construct the dataset $\mathcal{D}$ in Section \ref{sec:intervention_data}.

\section{Experimental Setup}
In this section, we describe the data used to perform the interventions and to measure the causal effects.

\subsection{Datasets}
For our analyses, we use instances of math word problems from three popular datasets: ASDiv-A \cite{miao-etal-2020-diverse}, MAWPS \cite{koncel-kedziorski-etal-2016-mawps}, and SVAMP \cite{patel-etal-2021-nlp}. 
The examples contained in these collections are pairs $(\bm{t},\bm{o})$ consisting of a question template $\bm{t}$ with its annotated operations $\bm{o}$.
Each of these pairs can be instantiated multiple times into problems $\bm{q} = (\bm{t}, \bm{n})$ by filling the template with numerical values $(n_1, n_2, \dots)$ and computing the ground-truth result $g = f_{\bm{o}}(\bm{n})$ (most problems involve two to three operands, i.e., $|\bm{n}| \in \{2, 3\}$).
We select a set of 437 two-operand and 307 three-operand template-expression pairs that we use to generate pairs of prompts representing an intervention. More details about the prompt generation procedure are in Appendix \ref{appendix:prompt_creation}. 
We use $(\bm{t}, \bm{n})$ to refer to an instantiated template that we use as a prompt.

\subsection{Intervention Data}
\label{sec:intervention_data}

Given an MWP $\bm{q} = (\bm{t}, \bm{n})$ and its solution $g$, we generate a second problem-solution instance $(\bm{q}', g')$
depending on the type of causal effect $\mathrm{CE}$ we want to measure and on the considered variable.
When intervening on the operands of the problem, the text of the problem is kept unaltered and a set of new operands $\bm{n}$ is sampled in such a way that the result $g$ is affected or not depending on the effect that is being measured. 
When changing the textual description of the problem, we change $\bm{t}$ such that either $\bm{o}' = \bm{o}$, or $\bm{o}' \neq \bm{o}$. In the former case, we sample a different template $\bm{t}' = (s', \bm{o})$ from the set of templates describing the same operations $\bm{o}$, in the latter case we sample a new $\bm{t}'$ describing a different operation. In Appendix \ref{appendix:examples} we report some examples of $(\bm{q}, \bm{q}')$ pairs representing the different types of interventions.

Given a model, we use the question pair $(\bm{q}, \bm{q}')$ to obtain a pair of answer distributions $\mathbb{P}(R| \bm{t}, \bm{n})$ and $\mathbb{P}(R| \bm{t}', \bm{n}')$, which we use to measure the causal effect of the intervention. 
We consider the space for the numerical values to be $\mathcal{I} = \{1, 2, \dots, C\}$ consisting of integer values, following the setup of several existing MWP datasets \cite{miao-etal-2020-diverse,koncel-kedziorski-etal-2016-mawps,patel-etal-2021-nlp}.
To control our experimental costs and make sure the models keep the number as one token, we set $C=300$.
From all the tokens in a model's vocabulary, we focus on the probability assigned to the numbers in our numerical space $\mathcal{I}$, and thus we use $\mathbb{P}(R=r)$ to denote the normalized probability $\mathbb{P}_{\mathrm{raw}}(R=r)/Z$, where $Z=\sum_{r=1}^{C} \mathbb{P}_{\mathrm{raw}}(R=r)$, and $\mathbb{P}_{\mathrm{raw}}(x)$ is the raw probability score assigned to the vocabulary token $x$.
For each intervention type, we generate a dataset $\mathcal{D}$ consisting of $ (\bm{q}, \bm{q}')$ pairs.
Unless otherwise specified, for our experiments we generate 500 intervention pairs for each template, and results are averaged over three seeds.

\subsection{Models to Evaluate}
We use our framework to assess the robustness of reasoning in thirteen pre-trained language models. We consider five sizes of the GPT-2 model \cite{radford2019language}: distilled \cite{sanh2019distilbert}, small, medium, large, and XL. We evaluate four models from EleutherAI that were pre-trained on the Pile \cite{gao2020pile}: GPT-Neo 1.3B and 2.7B \cite{gpt-neo}, GPT-J-6B \cite{gpt-j}, and GPT-NeoX-20B \cite{black2022gpt}. We use HuggingFace Transformers \cite{wolf2019transformers}
to access the models.
Additionally, we experiment with a set of instruction-tuned versions of GPT-3 \cite{gpt3}: Instruct \cite{ouyang2022instructGPT}, Curie, Davinci-002, and Davinci-003.\footnote{The OpenAI ids for these models are, respectively,  \texttt{davinci-instruct-beta}, \texttt{text-curie-001}, \texttt{text-davinci-002}, and \texttt{text-davinci-003}.}
Experiments with GPT-3 are carried out under the constraints set by the OpenAI APIs\footnote{\url{https://openai.com/api/}}, which prevent us from computing the causal effect using the same procedure as for the other models. We report the details about how the metrics were computed for GPT-3 in Appendix \ref{appendix:gpt3_approx}.
In the reported results, we indicate with an asterisk ($^*$) the metrics that were influenced by this limitation.

\section{Results}
Our analyses focus primarily on two-operand problems (Sections \ref{sec:n_on_r_2ops} and \ref{sec:t_on_r_2ops}) and later extend to more complex problems that involve three operands (Section \ref{sec:3_ops}) for the models that perform best on the two-operand test bed.
We compare the direct causal effect $\mathrm{DCE}$ and the total causal effect $\mathrm{TCE}$ of $\bm{N}$ and $\bm{T}$ on $R$.  $\mathrm{DCE}$ represents the undesired effect for a model to being mistakenly responsive to a change in $\bm{N}$ or $\bm{T}$ not leading to a change in the result $g$ (low robustness), whereas higher values of  $\mathrm{TCE}$ indicate a higher ability of the model to correctly adjust the probability weight assigned to the new solution $g'$ after the intervention (high sensitivity).

\subsection{Effect of $\bm{N}$ on $R$}
\label{sec:n_on_r_2ops}
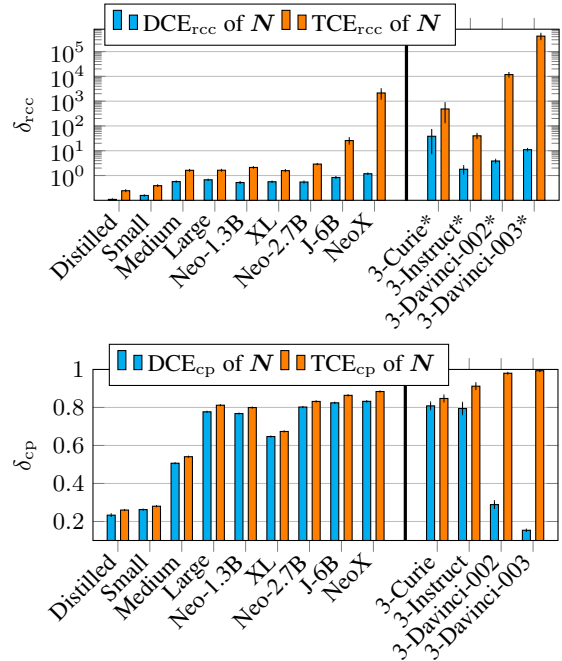
\begin{figure}\footnotesize
\begin{tikzpicture}
\begin{semilogyaxis}
[
	xtick=data,
	ymax=800000,
	ymin=0.1,
	log origin y=infty,
	ylabel=$\delta_{\mathrm{rcc}}$,
    ytick={1,10,100,1000,10000,100000,1000000},
	height=0.5\columnwidth,
	width=\columnwidth,
	symbolic x coords={Distilled, Small, Medium, Large, Neo-1.3B, XL, Neo-2.7B, J-6B, NeoX, \ , 3-Curie*, 3-Instruct*, 3-Davinci-002*, 3-Davinci-003*, A},
	enlarge y limits=0.0,
	enlarge x limits=0.06,
	legend style={at={(0.4,1.15)},
	anchor=north,legend columns=-1},
	ybar,
	bar width=3pt,
	grid=major,
    xmajorgrids=false,
    x tick label style={rotate=45,anchor=east}
]
\addplot+ [
   black, fill=cyan,
   error bars/.cd,
   y dir=both,
   y explicit,
   error mark options={
      rotate=90,
      mark size=0pt,
    }
]coordinates {
(J-6B, 0.834235346744347) +- (0, 0.04)
(Neo-1.3B, 0.522155872100273) +- (0, 0.027)
(Neo-2.7B, 0.546945810291668) +- (0, 0.024)
(Distilled, 0.108832419634291) +- (0, 0.001)
(Small, 0.157099819813293) +- (0, 0.00132896355741316)
(Large, 0.671930790279737) +- (0, 0.01)
(Medium, 0.569471490478381) +- (0, 0.01)
(XL, 0.56) +- (0, 0.00976850871032544)
(3-Curie*, 38.37348145494138) +- (0, 30.483747647342565)
(NeoX, 1.17916686887043) +- (0, 0.0278626018160753)
(3-Davinci-002*,3.8603336668546295) +- (0, 0.47435185634652777)
(3-Instruct*, 1.7914718574180748) +- (0, 0.5891337864978382)
(3-Davinci-003*, 11.025906942004847) +- (0, 0.8210212394041703)
};

\addplot+ [
   black, fill=orange,
   error bars/.cd,
   y dir=both,
   y explicit,
    error mark options={
      rotate=90,
      mark size=0pt,
    }
]coordinates {
(J-6B, 25.780298441897347) +- (0, 6.362351613767966)
(Neo-1.3B, 2.1197260504015434) +- (0, 0.04940484879745536)
(Neo-2.7B, 2.8916255381461853) +- (0, 0.03780760816893716)
(Distilled, 0.2434568953880113) +- (0, 0.00991243715811916)
(Small, 0.3892899513625348) +- (0, 0.011852482242094703)
(Large, 1.6354087671691948) +- (0, 0.06290616497617149)
(Medium, 1.6226962831730949) +- (0, 0.05577335022177161)
(XL, 1.5734177353962508) +- (0, 0.07869885155377497)
(3-Curie*, 485.15496502908155) +- (0, 344.38332170275)
(NeoX, 2140.875) +- (0, 885.048638522765)
(3-Davinci-002*, 11815.602815407865) +- (0, 1910.5459089278447)
(3-Davinci-003*, 421210.69134962215) +- (0, 103460.56199762033)
(3-Instruct*, 40.00695149530433) +- (0, 7.948463509136467)
};
\draw[very thick] (axis cs:\ ,0.001) -- (axis cs:\ ,10000000);,
    
\legend{DCE$_{\mathrm{rcc}}$ of $\bm{N}$, TCE$_{\mathrm{rcc}}$ of $\bm{N}$}
\end{semilogyaxis}
\end{tikzpicture}

\begin{tikzpicture}
\begin{axis}
[
	xtick=data,
	ymax=1,
	ymin=0.1,
	log origin y=infty,
	ylabel=$\delta_{\mathrm{cp}}$,
	height=0.5\columnwidth,
	width=\columnwidth,
	symbolic x coords={Distilled, Small, Medium, Large, Neo-1.3B, XL, Neo-2.7B, J-6B, NeoX, \ , 3-Curie, 3-Instruct, 3-Davinci-002, 3-Davinci-003, A},
	enlarge y limits=0.0,
	enlarge x limits=0.06,
	legend style={at={(0.4,1.15)},
	anchor=north,legend columns=-1},
	ybar,
	bar width=3pt,
	grid=major,
    xmajorgrids=false,
    x tick label style={rotate=45,anchor=east}
]
\addplot+ [
   black, fill=cyan,
   error bars/.cd,
   y dir=both,
   y explicit,
   error mark options={
      rotate=90,
      mark size=0pt,
    }
]coordinates {
(J-6B, 0.824442410373761) +- (0, 0.000732354431409923)
(Neo-1.3B, 0.7679359267734553) +- (0, 0.00001)
(Neo-2.7B, 0.802183066361557) +- (0, 0.0005354691075515)
(Distilled, 0.23335469107551488) +- (0, 0.0043971439847412495)
(Small, 0.26252631578947366) +- (0, 0.00001)
(Large, 0.777116704805492) +- (0, 0.00001)
(Medium, 0.5065812356979406) +- (0, 0.00001)
(XL, 0.6470160183066361) +- (0, 0.00001)
(3-Davinci-002, 0.28878718535469106) +- (0, 0.01774992567234188)
(3-Curie,0.8083142639206713) +- (0, 0.017886417848300213)
(NeoX, 0.8322) +- (0, 0.000721110255092798)
(3-Instruct, 0.7944317315026698) +- (0, 0.030136816154239656)
(3-Davinci-003, 0.15385202135774215) +- (0, 0.0033735079157125912)
};
\draw[very thick] (axis cs:\ ,0.001) -- (axis cs:\ ,1.1);,

\addplot+ [
   black, fill=orange,
   error bars/.cd,
   y dir=both,
   y explicit,
  error mark options={
      rotate=90,
      mark size=0pt,
    }
]coordinates {
(J-6B, 0.863734553775744) +- (0, 0.000662631111777246)
(Neo-1.3B, 0.7994279176201373) +- (0, 0.0001)
(Neo-2.7B, 0.83173767209665) +- (0, 0.000480962268167874)
(Distilled, 0.26010983981693364) +- (0, 0.0001)
(Small, 0.28039816933638445) +- (0, 0.0001)
(Large, 0.8118215102974828) +- (0, 0.0001)
(Medium, 0.5408695652173913) +- (0, 0.0001)
(XL, 0.673720823798627) +- (0, 0.0001)
(3-Curie,0.8472921434019831) +- (0, 0.0157771244041857)
(NeoX, 0.8839) +- (0, 0.000360555127546399)
(3-Davinci-002, 0.980091533180778) +- (0, 0.0021547592512319727)
(3-Instruct, 0.9118993135011441) +- (0, 0.015484738829031626)
(3-Davinci-003, 0.9933638443935927) +- (0, 0.00225374320407236)
};

\legend{DCE$_{\mathrm{cp}}$ of $\bm{N}$, TCE$_{\mathrm{cp}}$ of $\bm{N}$}
\end{axis}
\end{tikzpicture}
\vspace{-5pt}
\caption{Comparison of $\mathrm{DCE}(\bm{N} \rightarrow R)$ and $\mathrm{TCE}(\bm{N} \text{ on } R)$. $^*$approx values, see Appendix \ref{appendix:gpt3_approx}.}
\label{plot:effect_of_n}
\vspace{-8pt}
\end{figure}

\begin{figure*}[t]
    \centering
    \includegraphics[width=0.66\columnwidth]{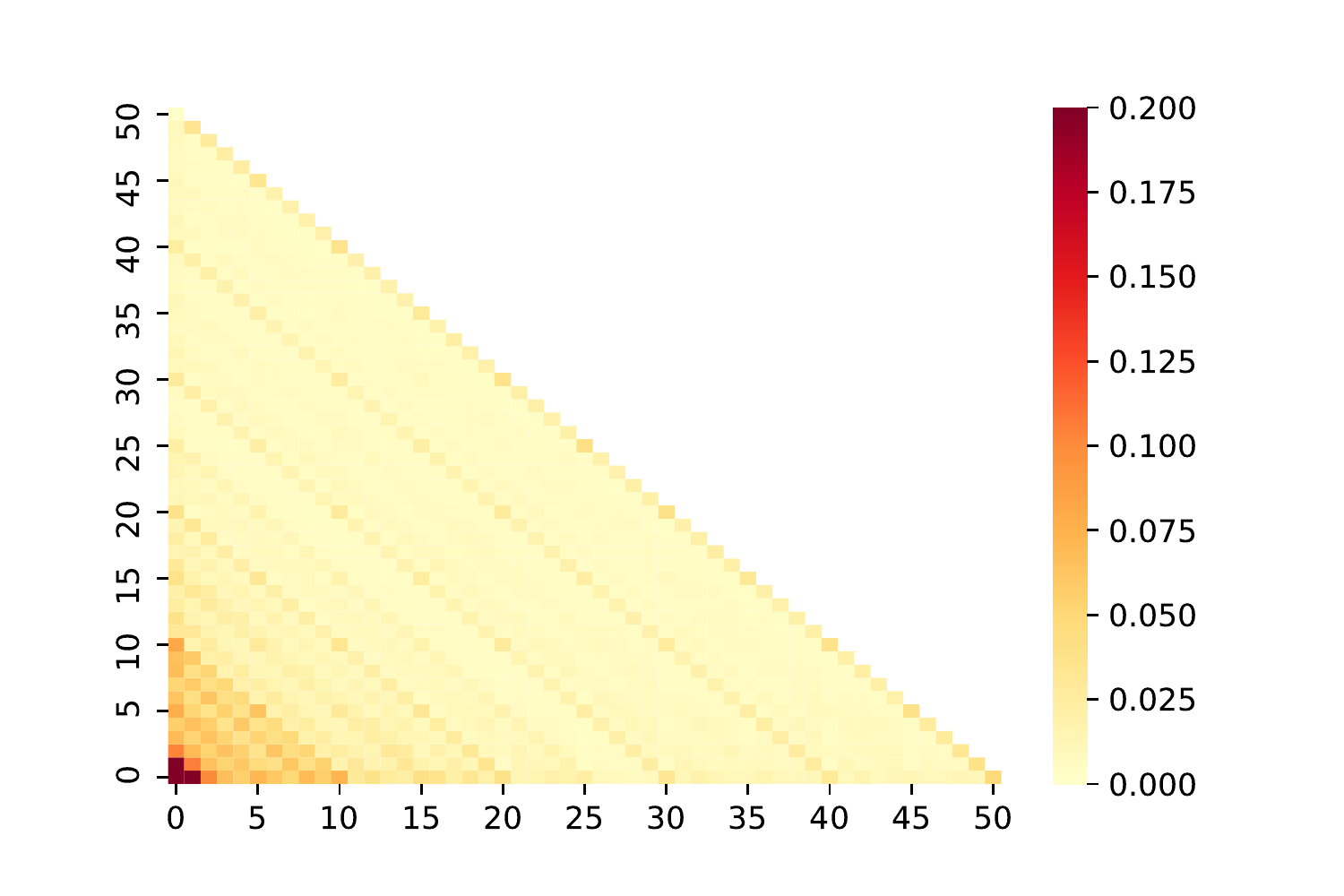}
    \includegraphics[width=0.66\columnwidth]{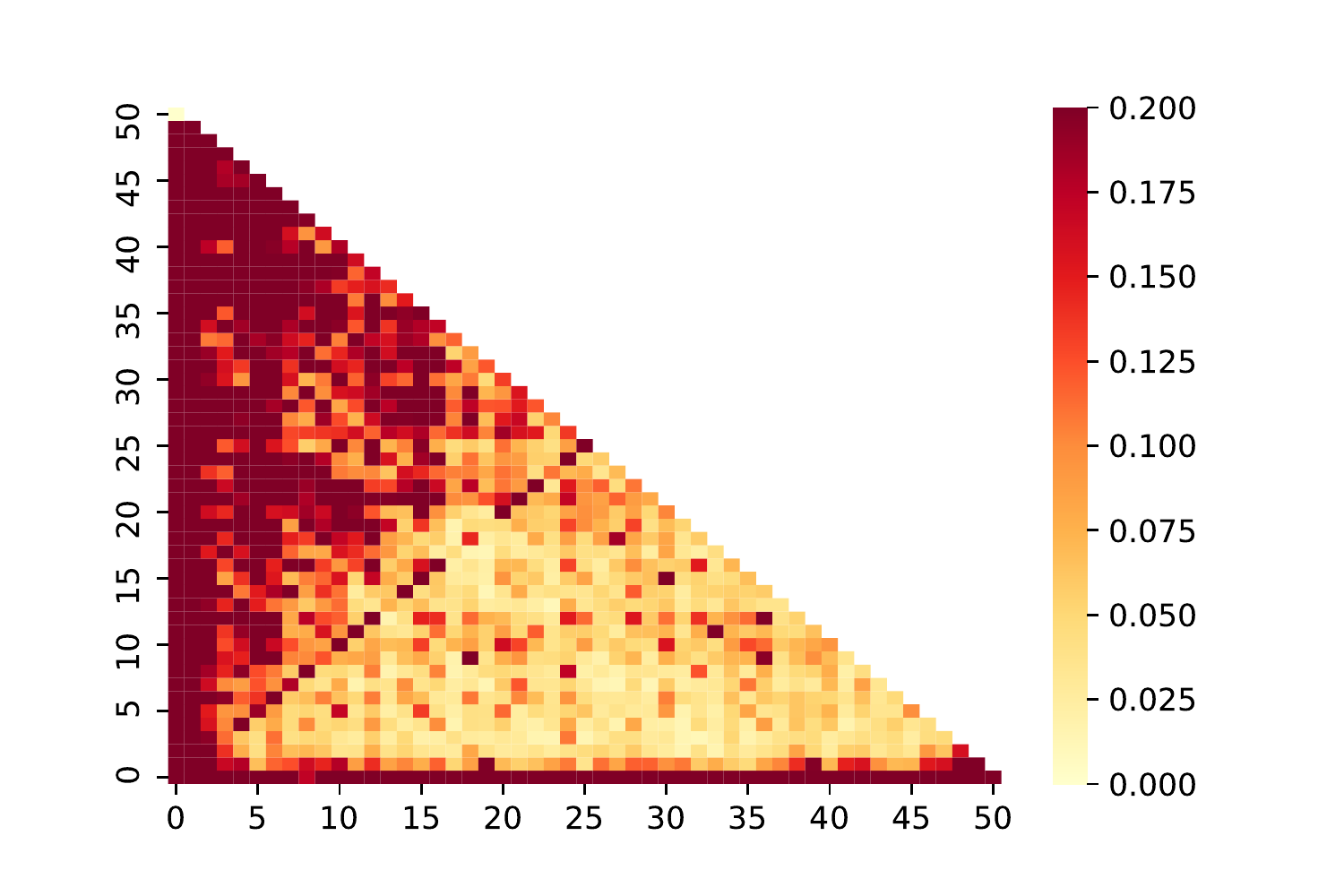}
    \includegraphics[width=0.66\columnwidth]{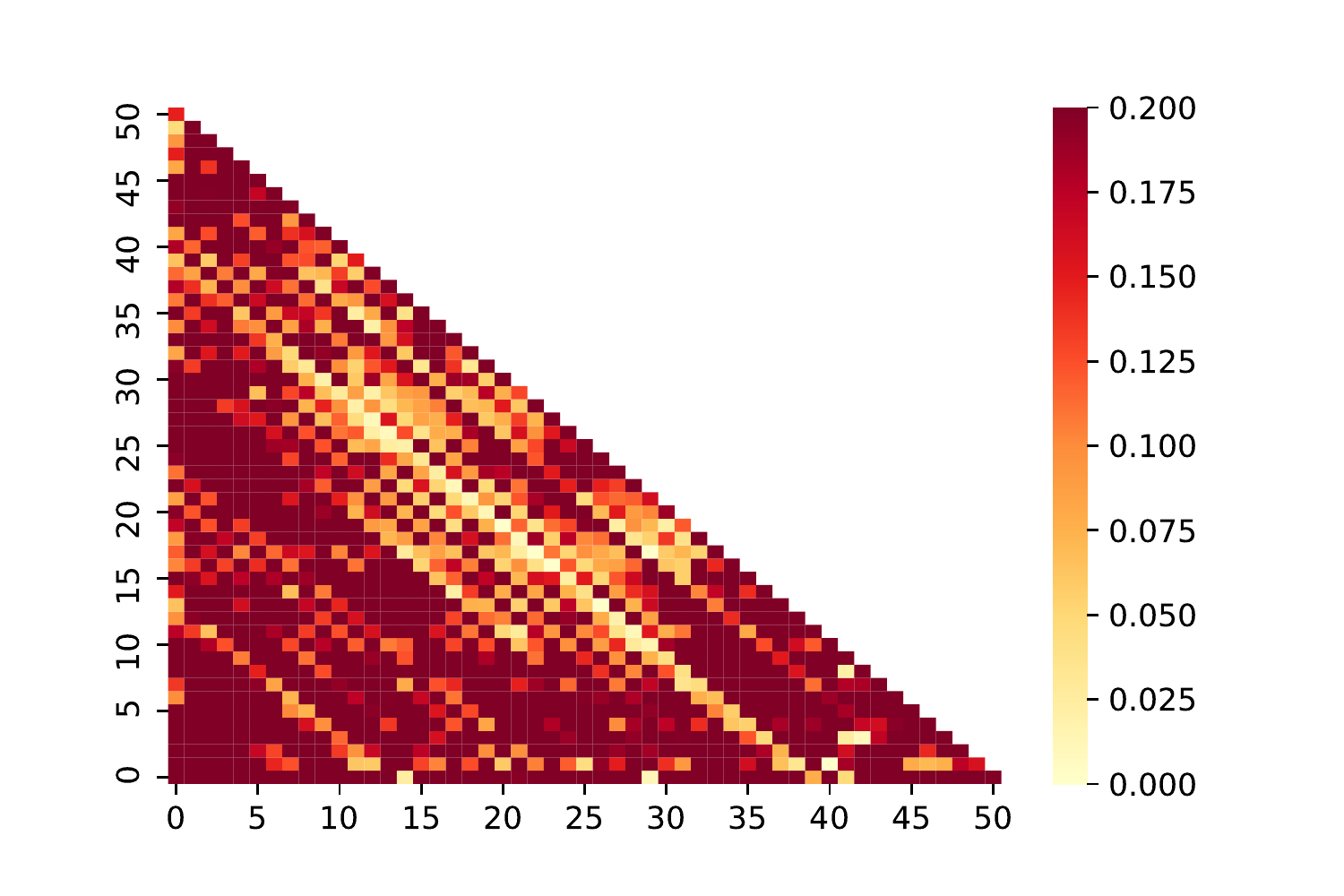}
    \caption{
    Heatmaps displaying $P(g)$ for Distil-GPT-2 (left), GPT-J-6B (center), and GPT-3 Davinci-002 (right). $g$ is the ground-truth result $g = n_1+n_2$ ($n_1$ and $n_2$ are represented by the x and y axes, respectively. The probability values for each combination of $((n_1, n_2), g)$ are averaged over 20 different templates.  Probability values over 0.2 are displayed with the darkest color.
    }
    \label{fig:heatmaps}
    \vspace{-10pt}
\end{figure*}

From the results in Figure \ref{plot:effect_of_n}, we notice that larger models exhibit a larger TCE$_{\mathrm{rcc}}$/DCE$_{\mathrm{rcc}}$ ratio.
In particular, in GPT-J-6B and NeoX, the TCE is, respectively, 30x and 1000x larger than the DCE.
However, this improvement in sensitivity is not manifested in terms of change of prediction ($\delta_{\mathrm{cp}}$), for which the models show to be affected by result-preserving changes almost as equally as by result-altering interventions. This behavior changes significantly in instruction-tuned models. In particular, for the 175B-parameter GPT-3, performance varies depending on the type of supervision, with the PPO-trained Davinci-003 exhibiting an 84\% difference between direct and total effect.

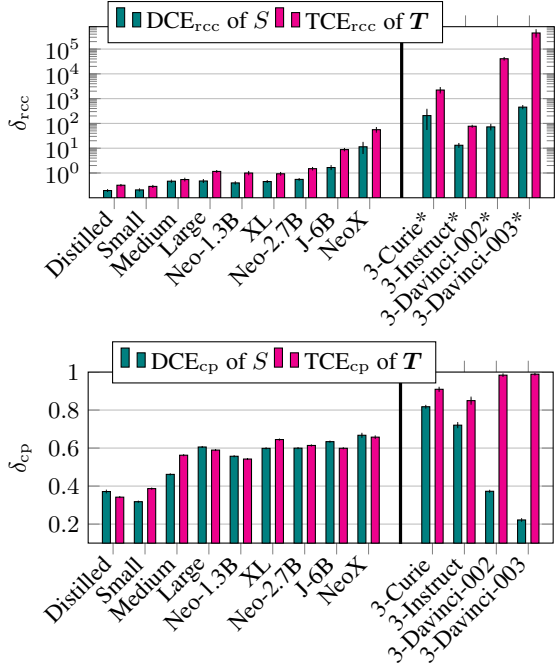
\begin{figure}[t]\footnotesize
\begin{tikzpicture}
\begin{semilogyaxis}[
	xtick=data,
	ymax=800000,
	ymin=0.1,
    ytick={1,10,100,1000,10000,100000,1000000},
	log origin y=infty,
	ylabel=$\delta_{\mathrm{rcc}}$,
	height=0.5\columnwidth,
	width=\columnwidth,
    symbolic x coords={Distilled, Small, Medium, Large, Neo-1.3B, XL, Neo-2.7B, J-6B, NeoX, \ , 3-Curie*, 3-Instruct*, 3-Davinci-002*, 3-Davinci-003*, A},
	enlarge y limits=0.0,
	enlarge x limits=0.06,
	legend style={at={(0.4,1.15)},
	anchor=north,legend columns=-1},
	ybar,
	bar width=3pt,
	grid=major,
    xmajorgrids=false,
    x tick label style={rotate=45,anchor=east}
]
\addplot+ [
   black, fill=teal,
   error bars/.cd,
   y dir=both,
   y explicit,
     error mark options={
      rotate=90,
      mark size=0pt,
    }
]coordinates {
(J-6B, 1.66141132299187) +- (0, 0.225)
(Neo-1.3B, 0.398066705758993) +- (0, 0.022)
(Neo-2.7B, 0.551559779439723) +- (0, 0.015)
(Distilled, 0.195896660986663) +- (0, 0.007)
(Small, 0.20761336577218) +- (0, 0.00971811540502479)
(Large, 0.47599176632657) +- (0, 0.0407676565497067)
(Medium, 0.466289443075747) +- (0, 0.0234487922448366)
(XL, 0.452318654151913) +- (0, 0.017891774509494)
(3-Curie*, 207.09022634500295) +- (0, 147.48511894560136)
(NeoX, 11.4182657266601) +- (0, 5)
(3-Davinci-002*,72.48066042572056) +- (0, 14.844075549654468)
(3-Instruct*, 13.150551053840516) +- (0, 1.6771769860754733)
(3-Davinci-003*, 454.4408496416752) +- (0, 44.605656629482915)

};

\addplot+ [
   black, fill=magenta,
   error bars/.cd,
   y dir=both,
   y explicit,
     error mark options={
      rotate=90,
      mark size=0pt,
    }
]coordinates {
(J-6B, 8.786903772873094) +- (0, 0.5434826780437797)
(Neo-1.3B, 0.9964369224026975) +- (0, 0.1060857609733839)
(Neo-2.7B, 1.4947411285112124) +- (0, 0.10159453344958379)
(Distilled, 0.32129470931913434) +- (0, 0.007003333999838406)
(Small, 0.2877384149650301) +- (0, 0.013008330346946696)
(Large, 1.1563198973806559) +- (0, 0.04193074654864333)
(Medium, 0.5445143435025176) +- (0, 0.041300853778617044)
(XL, 0.9287654038449459) +- (0, 0.07269771764576204)
(3-Curie*, 2223.413615634005) +- (0, 438.820275890894)
(NeoX, 55.8513333333333) +- (0, 8.94)
(3-Davinci-002*, 40710.92153893157) +- (0, 2746.9696297021583)
(3-Instruct*, 77.02657273094938) +- (0, 2.9102600975385986)
(3-Davinci-003*, 445878.50103203097) +- (0, 126506.5178705026)
};
\draw[very thick] (axis cs:\ ,0.0001) -- (axis cs:\ ,10000000);,
    
\legend{DCE$_{\mathrm{rcc}}$ of $S$, TCE$_{\mathrm{rcc}}$ of $\bm{T}$}
\end{semilogyaxis}
\end{tikzpicture}

\begin{tikzpicture}
\begin{axis}
[
	xtick=data,
	ymax=1,
	ymin=0.1,
	log origin y=infty,
	ylabel=$\delta_{\mathrm{cp}}$,
	height=0.5\columnwidth,
	width=\columnwidth,
	symbolic x coords={Distilled, Small, Medium, Large, Neo-1.3B, XL, Neo-2.7B, J-6B, NeoX, \ , 3-Curie, 3-Instruct, 3-Davinci-002, 3-Davinci-003, A},
	enlarge y limits=0.0,
	enlarge x limits=0.06,
	legend style={at={(0.4,1.18)},
	anchor=north,legend columns=-1},
	ybar,
	bar width=3pt,
	grid=major,
    xmajorgrids=false,
    x tick label style={rotate=45,anchor=east}
]
\addplot+ [
   black, fill=teal,
   error bars/.cd,
   y dir=both,
   y explicit,
     error mark options={
      rotate=90,
      mark size=0pt,
    }
]coordinates {
(J-6B, 0.6338293838862559) +- (0, 0.00001)
(Neo-1.3B, 0.556957345971564) +- (0, 0.00001)
(Neo-2.7B, 0.5996113744075829) +- (0, 0.00001)
(Distilled, 0.37133649289099524) +- (0, 0.0043971439847412495)
(Small, 0.318) +- (0, 0.00001)
(Large, 0.6056872037914692) +- (0, 0.00001)
(Medium, 0.46161137440758293) +- (0, 0.00001)
(XL, 0.5986445497630332) +- (0, 0.00001)
(3-Davinci-002, 0.3726698262243286) +- (0, 0.002576541300205386)
(3-Curie,0.8175355450236967) +- (0, 0.0045622524733629)
(NeoX,0.6674) +- (0,0.00763740793725201)
(3-Instruct, 0.7205371248025276) +- (0, 0.010558549643922253)
(3-Davinci-003, 0.2218009478672985) +- (0, 0.0037617317218927885)
};

\addplot+ [
   black, fill=magenta,
   error bars/.cd,
   y dir=both,
   y explicit,
  error mark options={
      rotate=90,
      mark size=0pt,
    }
]coordinates {
(J-6B, 0.5989649178255373) +- (0, 0.00001)
(Neo-1.3B, 0.5423119469026548) +- (0, 0.00001)
(Neo-2.7B, 0.6137701485461441) +- (0, 0.00001)
(Distilled, 0.3417845290771176) +- (0, 0.00001)
(Small, 0.38670393489254107) +- (0, 0.00001)
(Large, 0.5893844816687737) +- (0, 0.00001)
(Medium, 0.5624802465233881) +- (0, 0.00001)
(XL, 0.6448324905183312) +- (0, 0.00001)
(3-Davinci-002,.9837329633475319) +- (0, 0.0038282605239010337)
(3-Curie,0.9092859658113035) +- (0, 0.007838254348368172)
(NeoX,0.657566666666667) +- (0,0.00360185137579736)
(3-Instruct, 0.8496108305391257) +- (0, 0.015687273409255294)
(3-Davinci-003, 0.9888261976296984) +- (0, 0.001492852803797162)
};
\draw[very thick] (axis cs:\ ,0) -- (axis cs:\ ,1.1);,
    
\legend{DCE$_{\mathrm{cp}}$ of $S$, TCE$_{\mathrm{cp}}$ of $\bm{T}$}
\end{axis}
\end{tikzpicture}
\caption{Comparison of $\mathrm{DCE}(S \rightarrow R)$ and $\mathrm{TCE}(\bm{T} \text{ on } R)$. We use $^*$ to denote approximated values, explained in Appendix \ref{appendix:gpt3_approx}.}
\label{plot:effect_of_t}

\end{figure}

In Figure \ref{fig:heatmaps}, we present a different visualization of the direct causal effect of $\bm{N}$ on the model's prediction. We report the heatmaps showing the probability assigned by the model to the result $g$ of a problem $(\bm{t}, (n_1, n_2), g) \ | \ g = n_1 + n_2, \ \forall g \in \{0,1,\dots,50 \}, \ \forall (n_1, n_2) \in \{0,1,\dots,50 \}^2$. For Distil-GPT-2 we observe low overall probability assigned to $g$ and diagonal patterns indicating consistency in assigning higher probability to specific results (e.g., 10, 20, 30, 40, 50). For the two larger models we notice a higher probability mass assigned to the problem's result, but less consistency on the prediction of the same result with different sets of operands (this is true for GPT-J in particular). This result is consistent with the observed higher DCE and TCE in larger models: $P(g)$ might vary more considerably when intervening on $\bm{N}$ without affecting $g$, but overall the model assigns higher probability weight to the correct result, which correlates with higher sensitivity.

\subsection{Effect of $\bm{T}$ on $R$}
\label{sec:t_on_r_2ops}
In Figure \ref{plot:effect_of_t}, we report the total causal effect of the textual framing $\bm{T}$ and the direct causal effect of the irrelevant text elements $S$ on the model's prediction.
For the instruction-tuned models, the improvement in terms of prediction change ($\delta_{\mathrm{cp}}$) follows a similar trend as for $\bm{N}$, with GPT-3 Davinci-003 showing a 76\% difference between direct and total effect.
An interesting observation is that the irrelevant textual information $S$ appears to have a lower direct effect than $\bm{N}$ for all non-instruction-tuned models. However, in the GPT-3 Davinci-00x models, we observe the opposite (i.e., DCE$(\bm{N} \rightarrow R)$ $\leq$ DCE$(S \rightarrow R)$). This suggests that large instruction-based models tend to be more susceptible to variation in the textual framing of a problem, while smaller models are more responsive to changes in the numerical values (though not necessarily correctly).

\subsection{Overall Insights}
In comparison to other models, GPT-3 Davinci shows the highest DCE$_{\mathrm{rcc}}$, but low DCE$_{\mathrm{cp}}$. This discrepancy is related to the quantities that the two metrics consider. $\delta_{\mathrm{rcc}}$ takes into account the probability assigned to $g$, while $\delta_{\mathrm{cp}}$ does not consider the ground truth solution. One interpretation of this result is that GPT-3 Davinci consistently predicts the same answer $r = r'$ when $g = g'$, but the probabilities $P(g)$ and $P'(g)$ might vary significantly.

The results observed for the two kinds of intervention $\mathrm{do}(\bm{T}:\bm{t} \rightarrow \bm{t}')$ and $\mathrm{do}(\bm{N}:(n_1,n_2) \rightarrow (n_1', n_2'))$ show similar trends. Small models (Distilled and Small GPT-2) exhibit low sensitivity to interventions. Larger models (from GPT-2 Medium to GPT-Neo) appear to be more influenced by changes in both $\bm{N}$ and $\bm{T}$. However, they display similar sensitivity to both result-altering and result-preserving interventions. An improvement in sensitivity is noticeable in GPT-J and NeoX, though not accompanied by an improvement in robustness. Remarkably different behavior is instead shown by the GPT-3 Davinci models, which demonstrate substantially higher sensitivity to result-altering interventions (high TCE), and higher robustness (in terms of prediction change). 
In Appendix \ref{appendix:accuracy}, we report the accuracy of the models on the generated instances of MWPs, which exhibits a similar trend as the robustness/sensitivity changes we observed.

Possible explanations for the improved robustness and sensitivity demonstrated by the large GPT-3 models might be the dramatic size increase and extension/enhancement of the training procedure involving instructions.
The former idea is aligned with the \emph{emergent abilities} hypothesis \citep{wei2022emergent}, which postulates the existence of skills that are displayed by large-scale models but are not present in smaller-scale models. 
However, our observations show different performances in versions of GPT-3 Davinci that differ in the training procedure.\footnote{A high-level description of the training procedures for the models is provided at \url{https://beta.openai.com/docs/model-index-for-researchers}.}
This raises the question of whether the capability of LLMs to reason about math problems benefits from instruction-based tuning.
We address this question in the following section.

\subsection{Extending to LLaMA-Based Models}

To further investigate the roles played by size and training method in the model's performance, we carry out our experimental procedure on three versions with different sizes (7B, 13B, and 30B) of the LLaMA model \cite{touvron2023llama}, and on Stanford Alpaca (which applies instruction tuning on LLaMA 7B) \cite{alpaca}. 
We present these results separately, as the LLaMA tokenization makes the prediction setup different from the one used from the other models, and prevents us from computing the relative change in confidence ($\delta_{\mathrm{rcc}}$).\footnote{The LLaMA tokenizer considers each digit as an independent token in the vocabulary. This makes it problematic to compare the probability value assigned by the model to multi-digit numbers.}

\begin{figure}\footnotesize
\centering
\begin{tikzpicture}
\begin{axis}
[
	xtick=data,
	ymax=0.7,
	ymin=0.2,
    ytick={0.1,0.2,...,1},
	log origin y=infty,
	ylabel=$\delta_{\mathrm{cp}}$,
	height=0.5\columnwidth,
	width=0.75\columnwidth,
	symbolic x coords={\ LLaMA 7B\ \ , \ LLaMA 13B\ \ , \ LLaMA 30B\ \ , \ Alpaca \ },
	enlarge y limits=0.0,
	enlarge x limits=0.3,
	legend style={at={(0.5,1.15)},
	anchor=north,legend columns=-1},
	ybar,
	bar width=6pt,
	grid=major,
    xmajorgrids=false,
    x tick label style={rotate=45,anchor=east}
]
\addplot+ [
   black, fill=cyan,
   error bars/.cd,
   y dir=both,
   y explicit,
     error mark options={
      rotate=90,
      mark size=0pt,
    }
]coordinates {
(\ Alpaca \ , 0.2856) +- (0, 0.005338)
(\ LLaMA 7B\ \ , 0.3775) +- (0, 0.005338)
(\ LLaMA 13B\ \ , 0.3959) +- (0, 0.005231)
(\ LLaMA 30B\ \ , 0.3366) +- (0, 0.007149)
};
\addplot+ [
   black, fill=orange,
   error bars/.cd,
   y dir=both,
   y explicit,
     error mark options={
      rotate=90,
      mark size=0pt,
    }
]coordinates {
(\ Alpaca \ , 0.3801) +- (0, 0.005338)
(\ LLaMA 7B\ \ ,  0.4682) +- (0, 0.005185)
(\ LLaMA 13B\ \ , 0.6027) +- (0, 0.005234)
(\ LLaMA 30B\ \ , 0.5989) +- (0, 0.007415)
};
\legend{DCE of $\bm{N}$, TCE of $\bm{N}$}
\end{axis}
\end{tikzpicture}
\vspace{-5pt}
\caption{Comparison of direct and total effects of $N$ on $R$ for LLaMA and Alpaca.}
\label{plot:llama}
\vspace{-5pt}
\end{figure}
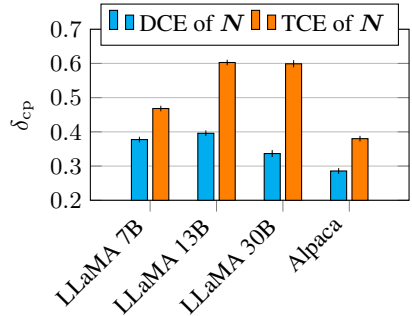

From the results (Figure \ref{plot:llama}), two notable observations emerge. Firstly, the increased difference between TCE and DCE observed with the increasing size of the LLaMA models suggests that a larger number of parameters can be a significant driver behind robustness/sensitivity improvement.
However, this is not necessarily the case across different models: GPT-NeoX-20B shows a smaller TCE$_{\mathrm{cp}}$-DCE$_{\mathrm{cp}}$ gap compared to LLaMA 7B (5.2\% vs 9.0\%).
Secondly, the instruction tuning procedure of Alpaca does not seem to help significantly with mathematical computation: the decrease in both TCE and DCE shows that robustness improves at the expense of sensitivity. Nonetheless, overall, when comparing Alpaca compared to its base model, LLaMA 7B, we observe an increase in the gap between TCE and DCE, although this difference is minimal (9.5\% vs 9.0\%).

The limited improvement of Alpaca might be attributed to its instruction tuning procedure consisting of ``a list of user-oriented instructions including email writing, social media, and productivity tools'' \cite{alpaca}, which differs from reasoning-intensive tasks.
We suggest future work to examine different types of instruction tuning (e.g., focused on reasoning procedures or reinforcement learning from human feedback), which might help the model answer more complex types of questions in a step-by-step manner and more accurately.
We hypothesize that the different performances in versions of GPT-3 Davinci might be produced by the specific type of instructions used for training, by the reinforcement learning component \cite{ouyang2022instructGPT}, or simply by an extension of the language modeling pre-training. It is challenging to pinpoint the exact factor in the training procedure that contributes to this improvement, as specific methodological details are not available.

\subsection{Moving to Three-Operand Problems}
\label{sec:3_ops}

\begin{figure}\footnotesize
\begin{tikzpicture}
\begin{semilogyaxis}
[
	xtick=data,
	ymax=12000,
	ymin=0.1,
	log origin y=infty,
	ylabel=$\delta_{\mathrm{rcc}}$,
    ytick={1,10,100,1000,10000, 100000, 1000000},
	height=0.5\columnwidth,
	width=0.5\columnwidth,
	symbolic x coords={3-Instruct*, 3-Davinci-002*, 3-Davinci-003*},
	enlarge y limits=0.0,
	enlarge x limits=0.3,
	legend style={at={(0.3,1.3)},
	anchor=north,legend columns=-1},
	ybar,
	bar width=5pt,
	grid=major,
    xmajorgrids=false,
    x tick label style={rotate=45,anchor=east}
]
\addplot+ [
   black, fill=cyan,
   error bars/.cd,
   y dir=both,
   y explicit,
     error mark options={
      rotate=90,
      mark size=0pt,
    }
]coordinates {
(3-Davinci-002*,35.73510970674531) +- (0, 11.576964545214814)
(3-Instruct*, 0.2654215096374149) +- (0, 0.06448478904336268)
(3-Davinci-003*, 204.42033846659515) +- (0, 112.14213366968319)
};

\addplot+ [
   black, fill=orange,
   error bars/.cd,
   y dir=both,
   y explicit,
     error mark options={
      rotate=90,
      mark size=0pt,
    }
]coordinates {
(3-Davinci-002*, 321.9426983017017) +- (0, 113.4233710415126)
(3-Davinci-003*, 4486.165919771869) +- (0, 440.25242681937243)
(3-Instruct*, 0.3811035764990454) +- (0, 0.22529671387767283)
};
\legend{DCE of $\bm{N}$, TCE of $\bm{N}$}
\end{semilogyaxis}
\end{tikzpicture}
\begin{tikzpicture}
\begin{axis}
[
	xtick=data,
	ymax=1,
	ymin=0.7,
	log origin y=infty,
	ylabel=$\delta_{\mathrm{cp}}$,
	height=0.5\columnwidth,
	width=0.5\columnwidth,
	symbolic x coords={\ 3-Instruct\ \ , \ 3-Davinci-002\ \ , \ 3-Davinci-003\ \ ,},
	enlarge y limits=0.0,
	enlarge x limits=0.3,
	legend style={at={(0.4,1.15)},
	anchor=north,legend columns=-1},
	ybar,
	bar width=5pt,
	grid=major,
    xmajorgrids=false,
    x tick label style={rotate=45,anchor=east}
]
\addplot+ [
   black, fill=cyan,
   error bars/.cd,
   y dir=both,
   y explicit,
     error mark options={
      rotate=90,
      mark size=0pt,
    }
]coordinates {
(\ 3-Instruct\ \ , 0.899457111834962) +- (0, 0.008925847519897764)
(\ 3-Davinci-003\ \ , 0.8764386536373507) +- (0, 0.029015632994062374)
(\ 3-Davinci-002\ \ , 0.8984799131378937) +- (0, 0.017821952658976598)
};
\addplot+ [
   black, fill=orange,
   error bars/.cd,
   y dir=both,
   y explicit,
     error mark options={
      rotate=90,
      mark size=0pt,
    }
]coordinates {
(\ 3-Davinci-002\ \ , 0.955157437567861) +- (0, 0.009145047422039358)
(\ 3-Instruct\ \ , 0.9072747014115091) +- (0, 0.011171366876779134)
(\ 3-Davinci-003\ \ , 0.9715526601520087) +- (0, 0.005996395774803076)
};
\end{axis}
\end{tikzpicture}
\vspace{-5pt}
\caption{Comparison of direct and total effects of $\bm{N}$ on $R$ for three-operand problems.}
\label{plot:effect_of_n_3ops}
\vspace{-5pt}
\end{figure}
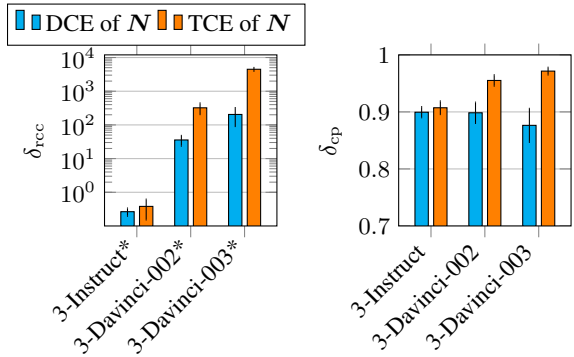

We extend our evaluation to consider the three-operand problems in the dataset. In these experiments, we consider only the GPT-3 175B-parameter models, as they are the only models performing well on the simpler bivariate problems. The results regarding the effects of $\bm{N}$ are reported in Figure \ref{plot:effect_of_n_3ops}.
We notice that the large difference between the desired (TCE) and undesired (DCE) effects observed on simpler problems shrinks significantly for both metrics. In particular, for Davinci-003, the direct effect of $\bm{N}$ (measured as $\delta_{\mathrm{cp}}$) grows from 0.17 to 0.87. That is, GPT-3 Davinci-003 predicts a different result 87\% of the time after an intervention that does not affect the ground-truth solution.
The increase in direct effect indicates a performance degradation in terms of brittleness: even the models that show good performance on two-operand problems, now display an unstable behavior after result-preserving interventions.

\section{Related Work}

\myparagraph{Causal NLP}
Causal inference aims to study the cause and effect from observational and interventional data \cite{pearl2009causality,peters2017elements}. Traditionally, researchers usually apply causal techniques to phenomena in nature and human society.
With the rise of powerful models in NLP, recent research has started to explore the intersection of causal inference and NLP, forming the study of Causal NLP \cite{jin-etal-2022-causalnlp,feder2021causal}. 

There are several formulations for Causal NLP: the \textit{causality for NLP} thread involves using the causal framework for data collection and task formulation \cite{jin-etal-2021-causal}, inspecting the (path-specific) causal effect of certain neurons on predictions \cite{vig2020investigating,meng2022locating}, understanding the causal effect of data and learning paradigm for model performance \cite{ni-etal-2022-original}, and as a way to frame prompts \citep{lyu2023psychologically}; and \textit{NLP for causality} involves testing the pure causal inference skills of LLMs \cite{jin2023causalbenchmark,jin2023large}, and use text as a variable for causal effect estimation \cite{roberts2020adjusting,veitch2020adapting,jin-etal-2021-mining-cause,jin2023ai}.

The most similar line of research to our work is the application of causal effect estimation on interpreting models' behavior, such as how models understand syntactic agreement \cite{finlayson-etal-2021-causal}, and how interventions in the representations and weights affect the model prediction \cite{feder-etal-2021-causalm}. 
To the best of our knowledge, our work is the first to formulate a causal framework for robustness behavioral tests, and also we are the first to introduce the idea to quantify the differences in the causal mechanisms of human reasoning and model decisions.

\myparagraph{Math Reasoning in NLP}
A growing body of work
tries to improve the math reasoning capability in NLP models \cite{zhang-etal-2020-language-embeddings, geva-etal-2020-injecting, spokoyny2021masked}, and prompting techniques for LLMs \citep{cobbe2021training,shen2021generate,kojima2022large,wei2022chain,chowdhery2022palm}. For analysis, significant attention has been given to models' ability to understand numerical quantities \cite{wallace-etal-2019-nlp, thawani-etal-2021-representing} and numerical operations \cite{pal-baral-2021-investigating-numeracy, berg-kirkpatrick-spokoyny-2020-empirical, piekos-etal-2021-measuring, razeghi2022impact}. 

\section{Conclusion}
We developed a framework to disentangle and separately measure the effect of different factors influencing the predictions of LLMs for math reasoning. 
Our results indicate that a drastic increase in both robustness and sensitivity emerges in the GPT-3 Davinci models. Additionally, we study the contribution of  size and instruction tuning in the models of the LLaMA family, observing that the Alpaca instruction tuning, while increasing the model's robustness, does not significantly improve the overall performance.
Our framework provides a formalized theory of behavioral testing for math reasoning models and opens new future directions to design behavioral tests of models in a principled way.

\section*{Ethical Considerations}
As for the ethical practice in this work,
the data involved are from existing MWP datasets with no private user information, and available under the MIT license. 
As for the ethical impact of the use of this work, the study is about providing a metric and analyzing existing models' robustness, so there is less concern over harmful usage. Rather, it is more about putting checks on existing AI models and helping humans understand them better before use. Potential stakeholders that could benefit from this research include NLP researchers working on math models, practitioners working on various applications involving mathematical reasoning with text, and e-learning design.

\section*{Limitations}
A key limitation in our work is that LLMs might have seen these math problems. Our work theoretically assumes this is not the case. 
Another limitation is that for the sake of simplicity, our work makes some assumptions. For example, we assume all numbers in the range of integers 0 to $C=300$. This would not cover every MWP out there. And future work is needed to generalize our framework to other forms of MWPs.
In this work, we are also constrained by the limitations of the OpenAI policy on the GPT-3 API. This limits the number of perturbations we consider in this work as well as the accuracy with which we can estimate our causal distributions. %
Finally, our work is restricted to English, and extending it to other languages will require us to create an MWP dataset in that language.

\ifarxiv
\section*{Acknowledgments}
This material is based in part upon works supported by the German Federal Ministry of Education and Research (BMBF): Tübingen AI Center, FKZ: 01IS18039B; by the Machine Learning Cluster of Excellence, EXC number 2064/1 – Project number 390727645; 
by the John Templeton Foundation (grant \#61156); by a Responsible AI grant by the Haslerstiftung; and an ETH Grant
(ETH-19 21-1).
Alessandro Stolfo is supported by armasuisse Science and Technology through a CYD Doctoral Fellowship.
Zhijing Jin is supported by PhD fellowships from the Future of Life Institute and Open Philanthropy, as well as the travel support from ELISE (GA no 951847) for the ELLIS program. We also thank OpenAI Researcher Access Program for granting our team credits to their API.
\fi

\bibliography{sec/refs_acl,sec/refs_this_paper,sec/refs_causality,sec/refs_zhijing,sec/refs_ai_safety}

\begin{thebibliography}{59}
\expandafter\ifx\csname natexlab\endcsname\relax\def\natexlab#1{#1}\fi

\bibitem[{Berg-Kirkpatrick and
  Spokoyny(2020)}]{berg-kirkpatrick-spokoyny-2020-empirical}
Taylor Berg-Kirkpatrick and Daniel Spokoyny. 2020.
\newblock \href {https://doi.org/10.18653/v1/2020.emnlp-main.385} {An empirical
  investigation of contextualized number prediction}.
\newblock In \emph{Proceedings of the 2020 Conference on Empirical Methods in
  Natural Language Processing (EMNLP)}, pages 4754--4764, Online. Association
  for Computational Linguistics.

\bibitem[{Black et~al.(2022)Black, Biderman, Hallahan, Anthony, Gao, Golding,
  He, Leahy, McDonell, Phang et~al.}]{black2022gpt}
Sid Black, Stella Biderman, Eric Hallahan, Quentin Anthony, Leo Gao, Laurence
  Golding, Horace He, Connor Leahy, Kyle McDonell, Jason Phang, et~al. 2022.
\newblock \href {https://arxiv.org/abs/2204.06745} {{GPT-NeoX-20B}: {An}
  open-source autoregressive language model}.
\newblock \emph{arXiv preprint arXiv:2204.06745}.

\bibitem[{Black et~al.(2021)Black, Gao, Wang, Leahy, and Biderman}]{gpt-neo}
Sid Black, Leo Gao, Phil Wang, Connor Leahy, and Stella Biderman. 2021.
\newblock \href {https://doi.org/10.5281/zenodo.5297715} {{GPT-Neo}: {Large}
  scale autoregressive language modeling with mesh-tensorflow}.
\newblock If you use this software, please cite it using these metadata.

\bibitem[{Brannon(2005)}]{doi:10.1073/pnas.0500328102}
Elizabeth~M. Brannon. 2005.
\newblock \href {https://doi.org/10.1073/pnas.0500328102} {The independence of
  language and mathematical reasoning}.
\newblock \emph{Proceedings of the National Academy of Sciences},
  102(9):3177--3178.

\bibitem[{Brown et~al.(2020)Brown, Mann, Ryder, Subbiah, Kaplan, Dhariwal,
  Neelakantan, Shyam, Sastry, Askell, Agarwal, Herbert-Voss, Krueger, Henighan,
  Child, Ramesh, Ziegler, Wu, Winter, Hesse, Chen, Sigler, Litwin, Gray, Chess,
  Clark, Berner, McCandlish, Radford, Sutskever, and Amodei}]{gpt3}
Tom Brown, Benjamin Mann, Nick Ryder, Melanie Subbiah, Jared~D Kaplan, Prafulla
  Dhariwal, Arvind Neelakantan, Pranav Shyam, Girish Sastry, Amanda Askell,
  Sandhini Agarwal, Ariel Herbert-Voss, Gretchen Krueger, Tom Henighan, Rewon
  Child, Aditya Ramesh, Daniel Ziegler, Jeffrey Wu, Clemens Winter, Chris
  Hesse, Mark Chen, Eric Sigler, Mateusz Litwin, Scott Gray, Benjamin Chess,
  Jack Clark, Christopher Berner, Sam McCandlish, Alec Radford, Ilya Sutskever,
  and Dario Amodei. 2020.
\newblock \href
  {https://proceedings.neurips.cc/paper/2020/file/1457c0d6bfcb4967418bfb8ac142f64a-Paper.pdf}
  {Language models are few-shot learners}.
\newblock In \emph{Advances in Neural Information Processing Systems},
  volume~33, pages 1877--1901. Curran Associates, Inc.

\bibitem[{Chowdhery et~al.(2022)Chowdhery, Narang, Devlin, Bosma, Mishra,
  Roberts, Barham, Chung, Sutton, Gehrmann et~al.}]{chowdhery2022palm}
Aakanksha Chowdhery, Sharan Narang, Jacob Devlin, Maarten Bosma, Gaurav Mishra,
  Adam Roberts, Paul Barham, Hyung~Won Chung, Charles Sutton, Sebastian
  Gehrmann, et~al. 2022.
\newblock \href {https://arxiv.org/abs/2204.02311} {Palm: Scaling language
  modeling with pathways}.
\newblock \emph{arXiv preprint arXiv:2204.02311}.

\bibitem[{Cobbe et~al.(2021)Cobbe, Kosaraju, Bavarian, Hilton, Nakano, Hesse,
  and Schulman}]{cobbe2021training}
Karl Cobbe, Vineet Kosaraju, Mohammad Bavarian, Jacob Hilton, Reiichiro Nakano,
  Christopher Hesse, and John Schulman. 2021.
\newblock \href {https://arxiv.org/abs/2110.14168} {Training verifiers to solve
  math word problems}.
\newblock \emph{arXiv preprint arXiv:2110.14168}.

\bibitem[{Feder et~al.(2021{\natexlab{a}})Feder, Keith, Manzoor, Pryzant,
  Sridhar, Wood-Doughty, Eisenstein, Grimmer, Reichart, Roberts, n~M.~Stewart,
  Veitch, and Yang}]{feder2021causal}
Amir Feder, Katherine~A. Keith, Emaad Manzoor, Reid Pryzant, Dhanya Sridhar,
  Zach Wood-Doughty, Jacob Eisenstein, Justin Grimmer, Roi Reichart,
  Margaret~E. Roberts, Brando n~M.~Stewart, Victor Veitch, and Diyi Yang.
  2021{\natexlab{a}}.
\newblock \href {http://arxiv.org/abs/2109.00725} {Causal inference in natural
  language processing: {E}stimation, prediction, interpretation and beyond}.
\newblock \emph{CoRR}, abs/2109.00725.

\bibitem[{Feder et~al.(2021{\natexlab{b}})Feder, Oved, Shalit, and
  Reichart}]{feder-etal-2021-causalm}
Amir Feder, Nadav Oved, Uri Shalit, and Roi Reichart. 2021{\natexlab{b}}.
\newblock \href {https://doi.org/10.1162/coli_a_00404} {{C}ausa{LM}: Causal
  model explanation through counterfactual language models}.
\newblock \emph{Computational Linguistics}, 47(2):333--386.

\bibitem[{Finlayson et~al.(2021)Finlayson, Mueller, Gehrmann, Shieber, Linzen,
  and Belinkov}]{finlayson-etal-2021-causal}
Matthew Finlayson, Aaron Mueller, Sebastian Gehrmann, Stuart Shieber, Tal
  Linzen, and Yonatan Belinkov. 2021.
\newblock \href {https://doi.org/10.18653/v1/2021.acl-long.144} {Causal
  analysis of syntactic agreement mechanisms in neural language models}.
\newblock In \emph{Proceedings of the 59th Annual Meeting of the Association
  for Computational Linguistics and the 11th International Joint Conference on
  Natural Language Processing (Volume 1: Long Papers)}, pages 1828--1843,
  Online. Association for Computational Linguistics.

\bibitem[{Gao et~al.(2020)Gao, Biderman, Black, Golding, Hoppe, Foster, Phang,
  He, Thite, Nabeshima et~al.}]{gao2020pile}
Leo Gao, Stella Biderman, Sid Black, Laurence Golding, Travis Hoppe, Charles
  Foster, Jason Phang, Horace He, Anish Thite, Noa Nabeshima, et~al. 2020.
\newblock \href {https://arxiv.org/abs/2101.00027} {The pile: {An} 800gb
  dataset of diverse text for language modeling}.
\newblock \emph{arXiv preprint arXiv:2101.00027}.

\bibitem[{Geva et~al.(2020)Geva, Gupta, and Berant}]{geva-etal-2020-injecting}
Mor Geva, Ankit Gupta, and Jonathan Berant. 2020.
\newblock \href {https://doi.org/10.18653/v1/2020.acl-main.89} {Injecting
  numerical reasoning skills into language models}.
\newblock In \emph{Proceedings of the 58th Annual Meeting of the Association
  for Computational Linguistics}, pages 946--958, Online. Association for
  Computational Linguistics.

\bibitem[{Jin et~al.(2021{\natexlab{a}})Jin, Jiang, Wang, Liu, Wang, Ren, and
  Qu}]{jin2021numgpt}
Zhihua Jin, Xin Jiang, Xingbo Wang, Qun Liu, Yong Wang, Xiaozhe Ren, and Huamin
  Qu. 2021{\natexlab{a}}.
\newblock \href {https://arxiv.org/abs/2109.03137} {Numgpt: {Improving}
  numeracy ability of generative pre-trained models}.
\newblock \emph{arXiv preprint arXiv:2109.03137}.

\bibitem[{Jin et~al.(2023{\natexlab{a}})Jin, Chen, Leeb, Gresele, Kamal, Lyu,
  Blin, Adauto, Kleiman-Weiner, Sachan, and
  Schoelkopf}]{jin2023causalbenchmark}
Zhijing Jin, Yuen Chen, Felix Leeb, Luigi Gresele, Ojasv Kamal, Zhiheng Lyu,
  Kevin Blin, Fernando~Gonzalez Adauto, Max Kleiman-Weiner, Mrinmaya Sachan,
  and Bernhard Schoelkopf. 2023{\natexlab{a}}.
\newblock Cladder: Assessing causal reasoning in language models.

\bibitem[{Jin et~al.(2022)Jin, Feder, and Zhang}]{jin-etal-2022-causalnlp}
Zhijing Jin, Amir Feder, and Kun Zhang. 2022.
\newblock \href {https://aclanthology.org/2022.emnlp-tutorials.4}
  {{C}ausal{NLP} tutorial: {A}n introduction to causality for natural language
  processing}.
\newblock In \emph{Proceedings of the 2022 Conference on Empirical Methods in
  Natural Language Processing: Tutorial Abstracts}, pages 17--22, Abu Dubai,
  UAE. Association for Computational Linguistics.

\bibitem[{Jin et~al.(2023{\natexlab{b}})Jin, Liu, Lyu, Poff, Sachan, Mihalcea,
  Diab, and Schoelkopf}]{jin2023large}
Zhijing Jin, Jiarui Liu, Zhiheng Lyu, Spencer Poff, Mrinmaya Sachan, Rada
  Mihalcea, Mona Diab, and Bernhard Schoelkopf. 2023{\natexlab{b}}.
\newblock Can large language models infer causation from correlation?

\bibitem[{Jin et~al.(2023{\natexlab{c}})Jin, Lyu, Ding, Sachan, Zhang,
  Mihalcea, and Schoelkopf}]{jin2023ai}
Zhijing Jin, Zhiheng Lyu, Yiwen Ding, Mrinmaya Sachan, Kun Zhang, Rada
  Mihalcea, and Bernhard Schoelkopf. 2023{\natexlab{c}}.
\newblock \href {https://zhijing-jin.com/files/papers/AIScholar_2023.pdf} {{AI
  Scholars: A} dataset for {NLP}-involved causal inference}.

\bibitem[{Jin et~al.(2021{\natexlab{b}})Jin, Peng, Vaidhya, Schoelkopf, and
  Mihalcea}]{jin-etal-2021-mining-cause}
Zhijing Jin, Zeyu Peng, Tejas Vaidhya, Bernhard Schoelkopf, and Rada Mihalcea.
  2021{\natexlab{b}}.
\newblock \href {https://doi.org/10.18653/v1/2021.findings-emnlp.27} {Mining
  the cause of political decision-making from social media: {A} case study of
  {COVID}-19 policies across the {US} states}.
\newblock In \emph{Findings of the Association for Computational Linguistics:
  EMNLP 2021}, pages 288--301, Punta Cana, Dominican Republic. Association for
  Computational Linguistics.

\bibitem[{Jin et~al.(2021{\natexlab{c}})Jin, von K{\"u}gelgen, Ni, Vaidhya,
  Kaushal, Sachan, and Schoelkopf}]{jin-etal-2021-causal}
Zhijing Jin, Julius von K{\"u}gelgen, Jingwei Ni, Tejas Vaidhya, Ayush Kaushal,
  Mrinmaya Sachan, and Bernhard Schoelkopf. 2021{\natexlab{c}}.
\newblock \href {https://doi.org/10.18653/v1/2021.emnlp-main.748} {Causal
  direction of data collection matters: {I}mplications of causal and anticausal
  learning for {NLP}}.
\newblock In \emph{Proceedings of the 2021 Conference on Empirical Methods in
  Natural Language Processing}, pages 9499--9513, Online and Punta Cana,
  Dominican Republic. Association for Computational Linguistics.

\bibitem[{Kaushik et~al.(2020)Kaushik, Hovy, and Lipton}]{kaushik2020learning}
Divyansh Kaushik, Eduard~H. Hovy, and Zachary~Chase Lipton. 2020.
\newblock \href {https://openreview.net/forum?id=Sklgs0NFvr} {Learning the
  difference that makes {A} difference with counterfactually-augmented data}.
\newblock In \emph{8th International Conference on Learning Representations,
  {ICLR} 2020, Addis Ababa, Ethiopia, April 26-30, 2020}. OpenReview.net.

\bibitem[{Kojima et~al.(2022)Kojima, Gu, Reid, Matsuo, and
  Iwasawa}]{kojima2022large}
Takeshi Kojima, Shixiang~Shane Gu, Machel Reid, Yutaka Matsuo, and Yusuke
  Iwasawa. 2022.
\newblock Large language models are zero-shot reasoners.
\newblock \emph{arXiv preprint arXiv:2205.11916}.

\bibitem[{Koncel-Kedziorski et~al.(2016)Koncel-Kedziorski, Roy, Amini, Kushman,
  and Hajishirzi}]{koncel-kedziorski-etal-2016-mawps}
Rik Koncel-Kedziorski, Subhro Roy, Aida Amini, Nate Kushman, and Hannaneh
  Hajishirzi. 2016.
\newblock \href {https://doi.org/10.18653/v1/N16-1136} {{MAWPS}: A math word
  problem repository}.
\newblock In \emph{Proceedings of the 2016 Conference of the North {A}merican
  Chapter of the Association for Computational Linguistics: Human Language
  Technologies}, pages 1152--1157, San Diego, California. Association for
  Computational Linguistics.

\bibitem[{Lyu et~al.(2023)Lyu, Jin, Mattern, Mihalcea, Sachan, and
  Sch{\"{o}}lkopf}]{lyu2023psychologically}
Zhiheng Lyu, Zhijing Jin, Justus Mattern, Rada Mihalcea, Mrinmaya Sachan, and
  Bernhard Sch{\"{o}}lkopf. 2023.
\newblock \href {https://doi.org/10.48550/arXiv.2305.01764}
  {Psychologically-inspired causal prompts}.
\newblock \emph{CoRR}, abs/2305.01764.

\bibitem[{Meng et~al.(2022)Meng, Bau, Andonian, and
  Belinkov}]{meng2022locating}
Kevin Meng, David Bau, Alex Andonian, and Yonatan Belinkov. 2022.
\newblock \href
  {https://proceedings.neurips.cc/paper_files/paper/2022/file/6f1d43d5a82a37e89b0665b33bf3a182-Paper-Conference.pdf}
  {Locating and editing factual associations in {GPT}}.
\newblock \emph{Advances in Neural Information Processing Systems},
  35:17359--17372.

\bibitem[{Miao et~al.(2020)Miao, Liang, and Su}]{miao-etal-2020-diverse}
Shen-yun Miao, Chao-Chun Liang, and Keh-Yih Su. 2020.
\newblock \href {https://doi.org/10.18653/v1/2020.acl-main.92} {A diverse
  corpus for evaluating and developing {E}nglish math word problem solvers}.
\newblock In \emph{Proceedings of the 58th Annual Meeting of the Association
  for Computational Linguistics}, pages 975--984, Online. Association for
  Computational Linguistics.

\bibitem[{Mishra et~al.(2022{\natexlab{a}})Mishra, Finlayson, Lu, Tang,
  Welleck, Baral, Rajpurohit, Tafjord, Sabharwal, Clark
  et~al.}]{mishra2022lila}
Swaroop Mishra, Matthew Finlayson, Pan Lu, Leonard Tang, Sean Welleck, Chitta
  Baral, Tanmay Rajpurohit, Oyvind Tafjord, Ashish Sabharwal, Peter Clark,
  et~al. 2022{\natexlab{a}}.
\newblock \href {URL_DOI_https://arxiv.org/abs/2210.17517MISSING} {Lila: {A}
  unified benchmark for mathematical reasoning}.
\newblock \emph{arXiv preprint arXiv:2210.17517}.

\bibitem[{Mishra et~al.(2022{\natexlab{b}})Mishra, Mitra, Varshney, Sachdeva,
  Clark, Baral, and Kalyan}]{mishra-etal-2022-numglue}
Swaroop Mishra, Arindam Mitra, Neeraj Varshney, Bhavdeep Sachdeva, Peter Clark,
  Chitta Baral, and Ashwin Kalyan. 2022{\natexlab{b}}.
\newblock \href {https://doi.org/10.18653/v1/2022.acl-long.246} {{N}um{GLUE}:
  {A} suite of fundamental yet challenging mathematical reasoning tasks}.
\newblock In \emph{Proceedings of the 60th Annual Meeting of the Association
  for Computational Linguistics (Volume 1: Long Papers)}, pages 3505--3523,
  Dublin, Ireland. Association for Computational Linguistics.

\bibitem[{Monti et~al.(2012)Monti, Parsons, and Osherson}]{monti2012thought}
Martin~M Monti, Lawrence~M Parsons, and Daniel~N Osherson. 2012.
\newblock \href
  {https://journals.sagepub.com/doi/abs/10.1177/0956797612437427?journalCode=pssa}
  {Thought beyond language: {Neural} dissociation of algebra and natural
  language}.
\newblock \emph{Psychological science}, 23(8):914--922.

\bibitem[{Ni et~al.(2022)Ni, Jin, Freitag, Sachan, and
  Sch{\"o}lkopf}]{ni-etal-2022-original}
Jingwei Ni, Zhijing Jin, Markus Freitag, Mrinmaya Sachan, and Bernhard
  Sch{\"o}lkopf. 2022.
\newblock \href {https://doi.org/10.18653/v1/2022.naacl-main.389} {Original or
  translated? {A} causal analysis of the impact of translationese on machine
  translation performance}.
\newblock In \emph{Proceedings of the 2022 Conference of the North American
  Chapter of the Association for Computational Linguistics: Human Language
  Technologies}, pages 5303--5320, Seattle, United States. Association for
  Computational Linguistics.

\bibitem[{Ouyang et~al.(2022)Ouyang, Wu, Jiang, Almeida, Wainwright, Mishkin,
  Zhang, Agarwal, Slama, Ray, Schulman, Hilton, Kelton, Miller, Simens, Askell,
  Welinder, Christiano, Leike, and Lowe}]{ouyang2022instructGPT}
Long Ouyang, Jeff Wu, Xu~Jiang, Diogo Almeida, Carroll~L. Wainwright, Pamela
  Mishkin, Chong Zhang, Sandhini Agarwal, Katarina Slama, Alex Ray, John
  Schulman, Jacob Hilton, Fraser Kelton, Luke Miller, Maddie Simens, Amanda
  Askell, Peter Welinder, Paul~F. Christiano, Jan Leike, and Ryan Lowe. 2022.
\newblock \href {https://doi.org/10.48550/arXiv.2203.02155} {Training language
  models to follow instructions with human feedback}.
\newblock \emph{CoRR}, abs/2203.02155.

\bibitem[{Pal and Baral(2021)}]{pal-baral-2021-investigating-numeracy}
Kuntal~Kumar Pal and Chitta Baral. 2021.
\newblock \href {https://doi.org/10.18653/v1/2021.findings-emnlp.265}
  {Investigating numeracy learning ability of a text-to-text transfer model}.
\newblock In \emph{Findings of the Association for Computational Linguistics:
  EMNLP 2021}, pages 3095--3101, Punta Cana, Dominican Republic. Association
  for Computational Linguistics.

\bibitem[{Patel et~al.(2021)Patel, Bhattamishra, and
  Goyal}]{patel-etal-2021-nlp}
Arkil Patel, Satwik Bhattamishra, and Navin Goyal. 2021.
\newblock \href {https://doi.org/10.18653/v1/2021.naacl-main.168} {Are {NLP}
  models really able to solve simple math word problems?}
\newblock In \emph{Proceedings of the 2021 Conference of the North American
  Chapter of the Association for Computational Linguistics: Human Language
  Technologies}, pages 2080--2094, Online. Association for Computational
  Linguistics.

\bibitem[{Pearl(1995)}]{pearl1995causal}
Judea Pearl. 1995.
\newblock \href {https://www.jstor.org/stable/2337329} {Causal diagrams for
  empirical research}.
\newblock \emph{Biometrika}, 82(4):669--688.

\bibitem[{Pearl(2001)}]{pearl2001direct}
Judea Pearl. 2001.
\newblock \href
  {https://dslpitt.org/uai/displayArticleDetails.jsp?mmnu=1\\&smnu=2\\&article\\_id=126\\&proceeding\\_id=17}
  {Direct and indirect effects}.
\newblock In \emph{{UAI} '01: Proceedings of the 17th Conference in Uncertainty
  in Artificial Intelligence, University of Washington, Seattle, Washington,
  USA, August 2-5, 2001}, pages 411--420. Morgan Kaufmann.

\bibitem[{Pearl(2009)}]{pearl2009causality}
Judea Pearl. 2009.
\newblock \href
  {https://www.cambridge.org/core/books/causality/B0046844FAE10CBF274D4ACBDAEB5F5B}
  {\emph{Causality}}.
\newblock Cambridge University Press.

\bibitem[{Peters et~al.(2017)Peters, Janzing, and
  Sch{\"o}lkopf}]{peters2017elements}
Jonas Peters, Dominik Janzing, and Bernhard Sch{\"o}lkopf. 2017.
\newblock \href {https://mitpress.mit.edu/books/elements-causal-inference}
  {\emph{Elements of causal inference: {F}oundations and learning algorithms}}.
\newblock The MIT Press.

\bibitem[{Pi{\k{e}}kos et~al.(2021)Pi{\k{e}}kos, Malinowski, and
  Michalewski}]{piekos-etal-2021-measuring}
Piotr Pi{\k{e}}kos, Mateusz Malinowski, and Henryk Michalewski. 2021.
\newblock \href {https://doi.org/10.18653/v1/2021.acl-short.49} {Measuring and
  improving {BERT}{'}s mathematical abilities by predicting the order of
  reasoning.}
\newblock In \emph{Proceedings of the 59th Annual Meeting of the Association
  for Computational Linguistics and the 11th International Joint Conference on
  Natural Language Processing (Volume 2: Short Papers)}, pages 383--394,
  Online. Association for Computational Linguistics.

\bibitem[{Radford et~al.(2019)Radford, Wu, Child, Luan, Amodei, and
  Sutskever}]{radford2019language}
Alec Radford, Jeff Wu, Rewon Child, David Luan, Dario Amodei, and Ilya
  Sutskever. 2019.
\newblock \href
  {https://d4mucfpksywv.cloudfront.net/better-language-models/language_models_are_unsupervised_multitask_learners.pdf}
  {Language models are unsupervised multitask learners}.

\bibitem[{Razeghi et~al.(2022)Razeghi, Logan~IV, Gardner, and
  Singh}]{razeghi2022impact}
Yasaman Razeghi, Robert~L Logan~IV, Matt Gardner, and Sameer Singh. 2022.
\newblock \href {https://arxiv.org/abs/2202.07206} {Impact of pretraining term
  frequencies on few-shot reasoning}.
\newblock \emph{arXiv preprint arXiv:2202.07206}.

\bibitem[{Roberts et~al.(2020)Roberts, Stewart, and
  Nielsen}]{roberts2020adjusting}
Margaret~E Roberts, Brandon~M Stewart, and Richard~A Nielsen. 2020.
\newblock \href {http://www.mit.edu/\textasciitilde rnielsen/textmatching.pdf}
  {Adjusting for confounding with text matching}.
\newblock \emph{American Journal of Political Science}, 64(4):887--903.

\bibitem[{Sachan et~al.(2017)Sachan, Dubey, and Xing}]{sachan2017textbooks}
Mrinmaya Sachan, Kumar Dubey, and Eric Xing. 2017.
\newblock \href {https://aclanthology.org/D17-1081/} {From textbooks to
  knowledge: {A} case study in harvesting axiomatic knowledge from textbooks to
  solve geometry problems}.
\newblock In \emph{Proceedings of the 2017 Conference on Empirical Methods in
  Natural Language Processing}, pages 773--784.

\bibitem[{Sachan et~al.(2018)Sachan, Dubey, Mitchell, Roth, and
  Xing}]{sachan2018learning}
Mrinmaya Sachan, Kumar~Avinava Dubey, Tom~M Mitchell, Dan Roth, and Eric~P
  Xing. 2018.
\newblock \href
  {https://proceedings.neurips.cc/paper/2018/hash/ac627ab1ccbdb62ec96e702f07f6425b-Abstract.html}
  {Learning pipelines with limited data and domain knowledge: {A} study in
  parsing physics problems}.
\newblock \emph{Advances in Neural Information Processing Systems}, 31.

\bibitem[{Sachan and Xing(2017)}]{sachan2017learning}
Mrinmaya Sachan and Eric Xing. 2017.
\newblock \href {https://aclanthology.org/S17-1029/} {Learning to solve
  geometry problems from natural language demonstrations in textbooks}.
\newblock In \emph{Proceedings of the 6th Joint Conference on Lexical and
  Computational Semantics (* SEM 2017)}, pages 251--261.

\bibitem[{Sanh et~al.(2019)Sanh, Debut, Chaumond, and
  Wolf}]{sanh2019distilbert}
Victor Sanh, Lysandre Debut, Julien Chaumond, and Thomas Wolf. 2019.
\newblock \href {https://arxiv.org/abs/1910.01108} {{DistilBERT,} a distilled
  version of {BERT}: {Smaller,} faster, cheaper and lighter}.
\newblock In \emph{NeurIPS EMC$^2$ Workshop}.

\bibitem[{Seo et~al.(2015)Seo, Hajishirzi, Farhadi, Etzioni, and
  Malcolm}]{seo-etal-2015-solving}
Minjoon Seo, Hannaneh Hajishirzi, Ali Farhadi, Oren Etzioni, and Clint Malcolm.
  2015.
\newblock \href {https://doi.org/10.18653/v1/D15-1171} {Solving geometry
  problems: {Combining} text and diagram interpretation}.
\newblock In \emph{Proceedings of the 2015 Conference on Empirical Methods in
  Natural Language Processing}, pages 1466--1476, Lisbon, Portugal. Association
  for Computational Linguistics.

\bibitem[{Shen et~al.(2021)Shen, Yin, Li, Shang, Jiang, Zhang, and
  Liu}]{shen2021generate}
Jianhao Shen, Yichun Yin, Lin Li, Lifeng Shang, Xin Jiang, Ming Zhang, and Qun
  Liu. 2021.
\newblock \href {https://arxiv.org/abs/2109.03034} {Generate \& rank: A
  multi-task framework for math word problems}.
\newblock \emph{arXiv preprint arXiv:2109.03034}.

\bibitem[{Spokoyny et~al.(2021)Spokoyny, Lee, Jin, and
  Berg-Kirkpatrick}]{spokoyny2021masked}
Daniel Spokoyny, Ivan Lee, Zhao Jin, and Taylor Berg-Kirkpatrick. 2021.
\newblock \href {https://arxiv.org/abs/2112.08616} {Masked measurement
  prediction: {Learning} to jointly predict quantities and units from textual
  context}.
\newblock \emph{arXiv preprint arXiv:2112.08616}.

\bibitem[{Taori et~al.(2023)Taori, Gulrajani, Zhang, Dubois, Li, Guestrin,
  Liang, and Hashimoto}]{alpaca}
Rohan Taori, Ishaan Gulrajani, Tianyi Zhang, Yann Dubois, Xuechen Li, Carlos
  Guestrin, Percy Liang, and Tatsunori~B. Hashimoto. 2023.
\newblock Stanford alpaca: An instruction-following llama model.
\newblock \url{https://github.com/tatsu-lab/stanford_alpaca}.

\bibitem[{Thawani et~al.(2021)Thawani, Pujara, Ilievski, and
  Szekely}]{thawani-etal-2021-representing}
Avijit Thawani, Jay Pujara, Filip Ilievski, and Pedro Szekely. 2021.
\newblock \href {https://doi.org/10.18653/v1/2021.naacl-main.53} {Representing
  numbers in {NLP}: a survey and a vision}.
\newblock In \emph{Proceedings of the 2021 Conference of the North American
  Chapter of the Association for Computational Linguistics: Human Language
  Technologies}, pages 644--656, Online. Association for Computational
  Linguistics.

\bibitem[{Touvron et~al.(2023)Touvron, Lavril, Izacard, Martinet, Lachaux,
  Lacroix, Rozi{\`e}re, Goyal, Hambro, Azhar et~al.}]{touvron2023llama}
Hugo Touvron, Thibaut Lavril, Gautier Izacard, Xavier Martinet, Marie-Anne
  Lachaux, Timoth{\'e}e Lacroix, Baptiste Rozi{\`e}re, Naman Goyal, Eric
  Hambro, Faisal Azhar, et~al. 2023.
\newblock \href {https://arxiv.org/abs/2302.13971} {Llama: Open and efficient
  foundation language models}.
\newblock \emph{arXiv preprint arXiv:2302.13971}.

\bibitem[{Veitch et~al.(2020)Veitch, Sridhar, and Blei}]{veitch2020adapting}
Victor Veitch, Dhanya Sridhar, and David~M. Blei. 2020.
\newblock \href {http://proceedings.mlr.press/v124/veitch20a.html} {Adapting
  text embeddings for causal inference}.
\newblock In \emph{Proceedings of the Thirty-Sixth Conference on Uncertainty in
  Artificial Intelligence, {UAI} 2020, virtual online, August 3-6, 2020},
  volume 124 of \emph{Proceedings of Machine Learning Research}, pages
  919--928. {AUAI} Press.

\bibitem[{Vig et~al.(2020)Vig, Gehrmann, Belinkov, Qian, Nevo, Singer, and
  Shieber}]{vig2020investigating}
Jesse Vig, Sebastian Gehrmann, Yonatan Belinkov, Sharon Qian, Daniel Nevo,
  Yaron Singer, and Stuart Shieber. 2020.
\newblock \href
  {https://proceedings.neurips.cc/paper/2020/hash/92650b2e92217715fe312e6fa7b90d82-Abstract.html}
  {Investigating gender bias in language models using causal mediation
  analysis}.
\newblock \emph{Advances in Neural Information Processing Systems},
  33:12388--12401.

\bibitem[{Wallace et~al.(2019)Wallace, Wang, Li, Singh, and
  Gardner}]{wallace-etal-2019-nlp}
Eric Wallace, Yizhong Wang, Sujian Li, Sameer Singh, and Matt Gardner. 2019.
\newblock \href {https://doi.org/10.18653/v1/D19-1534} {Do {NLP} models know
  numbers? probing numeracy in embeddings}.
\newblock In \emph{Proceedings of the 2019 Conference on Empirical Methods in
  Natural Language Processing and the 9th International Joint Conference on
  Natural Language Processing (EMNLP-IJCNLP)}, pages 5307--5315, Hong Kong,
  China. Association for Computational Linguistics.

\bibitem[{Wang and Komatsuzaki(2021)}]{gpt-j}
Ben Wang and Aran Komatsuzaki. 2021.
\newblock \href {https://github.com/kingoflolz/mesh-transformer-jax}
  {{GPT-J-6B}: {A} 6 billion parameter autoregressive language model}.

\bibitem[{Wei et~al.(2022{\natexlab{a}})Wei, Tay, Bommasani, Raffel, Zoph,
  Borgeaud, Yogatama, Bosma, Zhou, Metzler et~al.}]{wei2022emergent}
Jason Wei, Yi~Tay, Rishi Bommasani, Colin Raffel, Barret Zoph, Sebastian
  Borgeaud, Dani Yogatama, Maarten Bosma, Denny Zhou, Donald Metzler, et~al.
  2022{\natexlab{a}}.
\newblock \href {https://arxiv.org/abs/2206.07682} {Emergent abilities of large
  language models}.
\newblock \emph{arXiv preprint arXiv:2206.07682}.

\bibitem[{Wei et~al.(2022{\natexlab{b}})Wei, Wang, Schuurmans, Bosma, Chi, Le,
  and Zhou}]{wei2022chain}
Jason Wei, Xuezhi Wang, Dale Schuurmans, Maarten Bosma, Ed~H. Chi, Quoc Le, and
  Denny Zhou. 2022{\natexlab{b}}.
\newblock \href {http://arxiv.org/abs/2201.11903} {Chain of thought prompting
  elicits reasoning in large language models}.
\newblock \emph{CoRR}, abs/2201.11903.

\bibitem[{Welleck et~al.(2022)Welleck, West, Cao, and
  Choi}]{welleck2022symbolic}
Sean Welleck, Peter West, Jize Cao, and Yejin Choi. 2022.
\newblock \href {https://ojs.aaai.org/index.php/AAAI/article/view/20841}
  {Symbolic brittleness in sequence models: on systematic generalization in
  symbolic mathematics}.
\newblock In \emph{Proceedings of the AAAI Conference on Artificial
  Intelligence}, volume~36, pages 8629--8637.

\bibitem[{Wolf et~al.(2019)Wolf, Debut, Sanh, Chaumond, Delangue, Moi, Cistac,
  Rault, Louf, Funtowicz, and Brew}]{wolf2019transformers}
Thomas Wolf, Lysandre Debut, Victor Sanh, Julien Chaumond, Clement Delangue,
  Anthony Moi, Pierric Cistac, Tim Rault, R'emi Louf, Morgan Funtowicz, and
  Jamie Brew. 2019.
\newblock \href {https://arxiv.org/abs/1910.03771} {{HuggingFace}'s
  transformers: State-of-the-art natural language processing}.
\newblock \emph{arXiv preprint arXiv:1910.03771}.

\bibitem[{Zhang et~al.(2020)Zhang, Ramachandran, Tenney, Elazar, and
  Roth}]{zhang-etal-2020-language-embeddings}
Xikun Zhang, Deepak Ramachandran, Ian Tenney, Yanai Elazar, and Dan Roth. 2020.
\newblock \href {https://doi.org/10.18653/v1/2020.findings-emnlp.439} {Do
  language embeddings capture scales?}
\newblock In \emph{Findings of the Association for Computational Linguistics:
  EMNLP 2020}, pages 4889--4896, Online. Association for Computational
  Linguistics.

\end{thebibliography}
\bibliographystyle{acl_natbib}

\appendix

\section{Creation of the Prompts}
\label{appendix:prompt_creation}

We consider MWP examples from the union of the three datasets SVAMP, ASDiv-A, and MAWPS. The textual template $\bm{t}$ of a problem consists of a context (describing a real-world state and/or actions) and a question. 
In order to obtain suitable prompts for the models, we convert the problems' questions into statements where the result of the problem is expected to be the first token after the prompt.
E.g., in the example in section \ref{sec:problem_setup}, \emph{how many trees will he have?} is converted into \emph{the number of trees that he will have is \_}.
From the MWP templates of the SVAMP/ASDiv-A/MAWPS collection (we consider all splits), we filter out the templates whose questions do not start with \emph{How many...}, and we use spaCy\footnote{\url{https://spacy.io}} to identify the subject, the object and the verbs in the sentence. This allows us to convert the last sentence of the template from \emph{The number of... is}. This way, we obtain 437 statement-based MWP templates for two-operand problems and 307 for three-operand problems. We manually checked a subset of the templates to identify possible mistakes in the conversion procedure.

\section{Frequently Asked Questions}

\subsection{How do the intervention data look like?}
\label{appendix:examples}
In Table \ref{table:AppendixExamples} we report examples of MWP pairs representing different types of intervention.

\subsection{What is the accuracy of the evaluated models on the generated problems?}
\label{appendix:accuracy}
\begin{figure}\small
\begin{tikzpicture}
\begin{axis}
[
	xtick=data,
	ymax=1,
	ymin=0,
	ylabel= Avg. Accuracy,
	height=0.5\columnwidth,
	width=\columnwidth,
	symbolic x coords={Distilled, Regular, Medium, Large, XL, Neo-1.3B, Neo-2.7B, J-6B, NeoX, \ , 3-Curie, Davinci-Instruct, 3-Davinci-002, 3-Davinci-003, A},
	enlarge y limits=0.0,
	enlarge x limits=0.1,
	legend style={at={(0.40,0.95)},
	anchor=north,legend columns=-1},
	ybar stacked,
	bar width=7pt,
	grid=major,
    xmajorgrids=false,
    x tick label style={rotate=45,anchor=east}
]
\addplot+ [
   black, fill=orange!50!red,
   error bars/.cd,
   y dir=both,
   y explicit,
        error mark options={
      rotate=90,
      mark size=0pt,
    }
]coordinates {
(3-Davinci-002, 0.6955377574370709) +- (-, 0.003546910755148736)
(J-6B, 0.015441647597254006) +- (0, 0.00045308924485125846)
(Neo-2.7B, 0.004979405034324943) +- (-, 9.153318077802838e-06)
(Neo-1.3B, 0.00274370709382151) +- (-, 0.00010297482837528594)
(XL, 0.009892448512585812) +- (-, 0.0007162471395881012)
(Large, 0.00417162471395881) +- (-, 0.0005697940503432495)
(Medium, 0.002446224256292906) +- (-, 0.00011670480549199084)
(Regular, 0.00497254004576659) +- (-, 0.0006292906178489701)
(Distilled, 0.002556064073226545) +- (-, 0.0007254004576659038)
(3-Curie, 0.03609839816933638) +- (-, 0.00326086956521739)
(NeoX, 0.04302974828375286) +- (-, 0.00105720823798627)
(Davinci-Instruct, 0.12368421052631579) +- (-, 0.020938215102974826)
(3-Davinci-003, 0.8271167048054919) +- (-, 0.022997711670480536)
};

\addplot+ [
   black, fill=orange!50!yellow,
   error bars/.cd,
   y dir=both,
   y explicit,
        error mark options={
      rotate=90,
      mark size=0pt,
    }
]coordinates {
(3-Davinci-002, 0.16956521739130437) +- (0, 0.0017162471395880674)
(J-6B, 0.21408237986270023) +- (0, 0.002155606407322652)
(Neo-2.7B, 0.15461098398169335) +- (0, 0.0007665903890160114)
(Neo-1.3B, 0.13646681922196793) +- (0, 0.0019244851258581191)
(XL, 0.13179176201373) +- (0, 0.00034782608695652084)
(Large, 0.11921510297482839) +- (0, 0.0011945080091533208)
(Medium, 0.11280778032036615) +- (0, 0.00027917620137300064)
(Regular, 0.13502974828375286) +- (0, 0.004995423340961105)
(Distilled, 0.12067963386727688) +- (0, 0.003290617848970251)
(3-Curie, 0.16802059496567506) +- (0, 0.0026315789473684292)
(NeoX, 0.24368878718535467) +- (0, 0.0010892448512585806)
(Davinci-Instruct, 0.3392448512585812) +- (0, 0.018535469107551494)
(3-Davinci-003, 0.09839816933638446) +- (0, 0.0008009153318077611)

};
\draw[very thick] (axis cs:\ ,0) -- (axis cs:\ ,1.2);,
    
\legend{Accuracy@1, Accuracy@10}
\end{axis}
\end{tikzpicture}

\begin{tikzpicture}
\begin{axis}
[
	xtick=data,
	ymax=0.05,
	ymin=0,
	ylabel= Avg. Accuracy,
	height=0.5\columnwidth,
	width=\columnwidth,
	symbolic x coords={Distilled, Regular, Medium, Large, XL, Neo-1.3B, Neo-2.7B, J-6B, NeoX, \ , 3-Curie, A},
	enlarge y limits=0.0,
	enlarge x limits=0.1,
	legend style={at={(0.25,0.95)},
	anchor=north,legend columns=-1},
	ybar stacked,
	bar width=7pt,
	grid=major,
    xmajorgrids=false,
    x tick label style={rotate=45,anchor=east}
]
\addplot+ [
   black, fill=orange!50!red,
   error bars/.cd,
   y dir=both,
   y explicit,
        error mark options={
      rotate=90,
      mark size=0pt,
    }
]coordinates {
(J-6B, 0.015441647597254006) +- (0, 0.00045308924485125846)
(Neo-2.7B, 0.004979405034324943) +- (-, 9.153318077802838e-06)
(Neo-1.3B, 0.00274370709382151) +- (-, 0.00010297482837528594)
(XL, 0.009892448512585812) +- (-, 0.0007162471395881012)
(Large, 0.00417162471395881) +- (-, 0.0005697940503432495)
(Medium, 0.002446224256292906) +- (-, 0.00011670480549199084)
(Regular, 0.00497254004576659) +- (-, 0.0006292906178489701)
(Distilled, 0.002556064073226545) +- (-, 0.0007254004576659038)
(3-Curie, 0.03609839816933638) +- (-, 0.00326086956521739)
(NeoX, 0.04302974828375286) +- (-, 0.00105720823798627)
};
\draw[very thick] (axis cs:\ ,0) -- (axis cs:\ ,1.1);,

\legend{Accuracy@1}
\end{axis}
\end{tikzpicture}

\caption{Average accuracy of the models on the generated instances of MWPs. Results are averaged over two sets consisting of 500 problem instances generated for each template. The lower figure shows a zoomed-in visualization of the accuracy at 1.}
\label{fig:accuracy}
\end{figure}
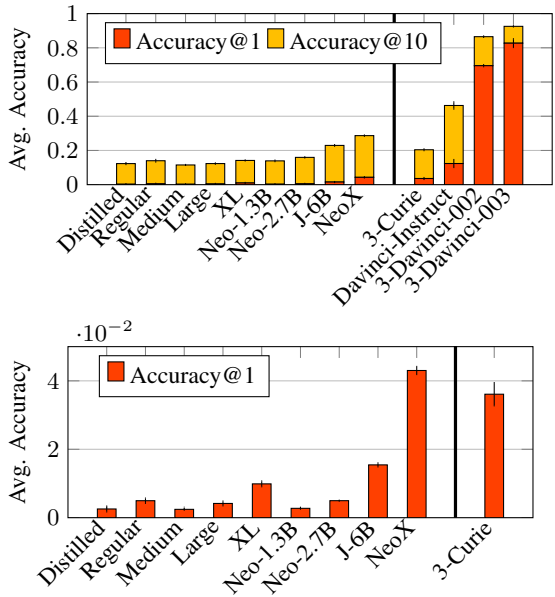

We report the accuracy of the models considered for evaluation in terms of accuracy at 1 and accuracy at 10. Results are displayed in Figure \ref{fig:accuracy}.

\begin{table*}
    \resizebox{\textwidth}{!}{
    \centering
    \begin{tabularx}{\textwidth}{ lX c}
    \toprule[0.1em]
    \multirow{4}{*}{$\mathrm{TCE}(\bm{N} \rightarrow R)$ } & Ruby has 87 candies. If she shares the candies among 29 friends, the number of candies that each friend gets is & \multirow{2}{*}{$g = 87 / 29 = 3$} \\
    \cmidrule{2-3}
    & Ruby has 35 candies. If she shares the candies among 5 friends, the number of candies that each friend gets is & \multirow{2}{*}{$g = 35 / 5 = 7$} \\
    \midrule[0.1em]
    \multirow{4}{*}{$\mathrm{DCE}(\bm{N} \rightarrow R)$ } & The school is composed of 13 buildings each having 10 classrooms. The number of classrooms that the school has is  & \multirow{2}{*}{$g = 10 \times 13 = 130$} \\
    \cmidrule{2-3}
    & The school is composed of 65 buildings each having 2 classrooms. The number of classrooms that the school has is & \multirow{2}{*}{$g = 65 \times 2 = 130$} \\
    \midrule[0.1em]
    \multirow{6}{*}{$\mathrm{DCE}(S \rightarrow R)$ } &  The razorback t-shirt shop ordered 6 cases of t-shirts. If each case contains 17 t-shirts the number of t-shirts that they ordered is  & \multirow{3}{*}{$g = 17 \times 6 = 102$} \\
    \cmidrule{2-3}
    & The roller coaster at the state fair costs 6 tickets per ride. If 17 friends were going to ride the roller coaster the number of tickets that they would need is  & \multirow{3}{*}{$g = 17 \times 6 = 102$} \\
    \midrule[0.1em]
    \multirow{5}{*}{$\mathrm{TCE}(\bm{T} \rightarrow R)$ } & Sean has 23 whistles. He has 6 more whistles than Charles. The number of whistles that Charles has is  & \multirow{2}{*}{$g = 23 - 6 = 17$} \\
    \cmidrule{2-3}
    &  Jovana filled her bucket with 23 pounds of shells. If she adds 6 more pounds of shell to fill her bucket, the number of pounds that she has is  & \multirow{3}{*}{$g = 23 + 6 = 29$} \\
   
 \bottomrule[0.1em]
\end{tabularx}
}
    \caption{For each of the causal effects measured (left column), we report a pair of MWPs illustrating the intervention performed (center), along with their respective ground-truth result (left column).}
    \label{table:AppendixExamples}
\end{table*}

\subsection{What is the relation between accuracy and the RCC metric?}

We examine the relationship between performance and robustness, computing the Pearson correlation coefficient between accuracy (accuracy@10) and the relative confidence change (RCC) metric. On a per-template basis (500 instances for each template), we found accuracy to be positively correlated with $\mathrm{TCE}(\bm{N} \text{ on } R)$ and $\mathrm{TCE}(T \text{ on } R)$  (0.24 and 0.49, respectively) and negatively correlated with $\mathrm{DCE}(\bm{N} \rightarrow R)$and $\mathrm{DCE}(S \rightarrow R)$ (-0.26 and -0.36, respectively). We see these results as a quantitative validation of the intuition behind our framework: the better the model’s performance, the more the model tends to correctly adjust its prediction after a result-altering intervention (higher sensitivity) and to correctly not change its prediction after a result-preserving intervention (higher robustness).

Moreover, we conduct an additional sanity check as in \citet{patel-etal-2021-nlp}: removing the question from the MWP templates, we observe a sensitivity-robustness degradation to random guessing (i.e., TCE $\simeq$ DCE). This indicates that the measurement of the causal effects within our framework is not affected by patterns in the templates that might have been picked up or memorized by large models.

\section{Computation of Causal Effects for GPT-3}
\label{appendix:gpt3_approx}
We access GPT-3 through the OpenAI APIs, which allow a user to prompt the model and obtain the probabilities assigned by the model to the $k$-th most likely vocabulary entries, for each token generated. 
To overcome this limitation, we approximate the relative probability change $\delta_{\mathrm{rcc}}$ as follows, depending on the kind of effect measured.

The limit for $k$ is set by OpenAI to 5. However, for our main set of experiments (i.e., computing the causal effects of $\bm{N}$, $S$, and $\bm{T}$) we were granted an increased limit of $k$ to 100. This allowed us to obtain reasonable estimates for the causal effects, as the number of cases in which $P(g)$ is not defined is less than $10\%$ of the number of examples that we consider.

\begin{algorithm}
\RestyleAlgo{ruled}
\caption{Computation of $\delta_{\mathrm{rcc}}$ for GPT-3}\label{alg}
$\bm{Q} = (\bm{t}, \bm{n}, g)$ \\
$\bm{Q}' = (\bm{t}', \bm{n}', g')$ \\
\eIf{$P(g)$ is defined}{
    \eIf{$P'(g)$ is defined}{
        $ \Delta = \frac{P(g) - P'(g)}{P'(g)}$ \\
    }{
        $\hat{P}' \gets P'(k\text{-th most likely token})$ \\
        $ \Delta = \frac{P(g) -\hat{P}'}{\hat{P}'}$ 
    }
}{
    $\Delta = 0$ 
}

\eIf{$P'(g')$ is defined}{
    \eIf{$P(g')$ is defined}{
        $ \Delta' = \frac{P'(g') - P(g')}{P(g')}$
    }{
        $\hat{P} \gets P(k\text{-th most likely token})$\\
        $ \Delta' = \frac{P'(g') -\hat{P}}{\hat{P}}$ \\
    }
}{
    $\Delta' = 0$ \\
}

$ \delta_{\mathrm{rcc}} = \frac{1}{2} (\Delta + \Delta')$

\end{algorithm}

\subsection{$\mathrm{TCE}(\bm{N} \text{ on } R)$ and $\mathrm{TCE}(\bm{T} \text{ on } R)$ }
In cases when $P(g)$ is defined (i.e. when $g$ appears in the top $k$ token predictions) and $P'(g)$ is not defined, we compute a lower bound on the relative change using the upper bound on $P'(g)$ given by the probability of the $k$-th most likely token. This gives us a conservative estimate of $\Delta$. For cases in which $P(g)$ is not defined, we cannot say anything about the relative change, and we set $\Delta = 0$. The same applies when swapping $P$ and $P'$. This procedure is illustrated by Algorithm \ref{alg}.

\subsection{$\mathrm{DCE}(\bm{N} \rightarrow R)$ and $\mathrm{DCE}(S \rightarrow R)$}
In this case, we simply discard the examples for which $P(g)$ is not defined or $P'(g)$ are not defined. In that is not the case, then we compute  $\delta_{\mathrm{rcc}}$ as in Section \ref{sec:distr_diff_metrics}.

\subsection{Heatmap Illustration}
The heatmap for GPT-3 displayed in Figure \ref{fig:heatmaps} was computed by taking the raw probability score produced by the model over the whole vocabulary, as the limit on the available top predicted tokens makes it impossible to normalize it over the set $\{0,\dots,300\}$, as done for the other models. The probability was set to 0 when $g$ did not appear in the model's top 5 predictions for the next token after the prompt.

\section{Computing Infrastructure \& Inference Details}
\label{appendix:computing_infrastructure}
To run our experiments, we used a single NVIDIA TITANRTX with 24GB of memory for all the versions of GPT-2 and GPT-Neo. We used a single NVIDIA A100 with 40GB of memory for GPT-J-6B and a single NVIDIA A100 with 80GB of memory for GPT-NeoX and the LLaMA models (two for the 30B version). We accessed GPT-3 using the OpenAI APIs. The longest run (GPT-J) on the four kinds of experiments corresponding to the four kinds of effects measured took $\sim$12 hours, using 500 MWP instances for each of the 437 templates. Due to budget and resource constraints, the experiments on GPT-3, GPT-NeoX, and LLaMA were carried out using 20 examples generated for each template and took $\sim$7 hours.
Experiment tracking was carried out using Weights \& Biases\footnote{\url{http://wandb.ai/}}.

\end{document}